\documentclass[11pt]{article}
\usepackage[a4paper,margin=1in]{geometry}
\usepackage{times}
\usepackage{amsmath,amssymb}
\usepackage{graphicx}
\usepackage{booktabs}
\usepackage{multirow}
\usepackage{placeins}
\usepackage{microtype}
\usepackage{url}
\usepackage{enumitem}
\usepackage{xcolor}
\usepackage[round,authoryear]{natbib}
\usepackage[colorlinks=true,allcolors=blue]{hyperref}
\setlist[itemize]{leftmargin=1.35em,itemsep=0.2em,topsep=0.25em}

\newcommand{\bs}{\textsc{BoundarySupport}}
\newcommand{\sg}{\operatorname{sg}}
\newcommand{\KCenter}{\operatorname{KCenter}}

\title{What Remains Normal?\\ Clean Images Miss Useful Near-Defect Normal Patches for Anomaly Detection}
\author{
Joongwon Chae$^{1,3}$ \qquad Runming Wang$^{1}$ \qquad Peiwu Qin$^{2}$\\
\normalfont
$^{1}$Institute of Biopharmaceutical and Health Engineering,\\
Tsinghua Shenzhen International Graduate School, Tsinghua University, Shenzhen, China\\
$^{2}$Guangdong Provincial Laboratory of Traditional Chinese Medicine,\\
Hengqin, Guangdong, China\\
$^{3}$RatelSoft
}
\date{}

\begin{document}
\maketitle

\begin{abstract}
Normal-only industrial anomaly detectors use patches from clean training images as normal references or reconstruction targets. This assumes that clean patches are sufficient for the normal regions encountered at test time. We test that assumption directly. On MVTec AD, admitting ground-truth-normal patches from real defect images to a DINOv2 memory candidate pool raises pixel average precision (P-AP) from 73.34 to 76.95 while keeping the encoder, test-time score, and number of stored references fixed. Patches within two patch cells of the annotated defect recover 94.70\% of this gain. We then ask whether useful patches of this kind can be exposed using clean training images alone. \bs{} inserts a procedural synthetic defect to alter surrounding context, excludes every token intersecting the nominal insertion or a detected RGB change, and learns only from pixel-preserved neighboring patches. Across three paired seeds, the same principle improves P-AP in all six memory/reconstruction--dataset settings across MVTec, VisA, and Real-IAD. Matched controls identify the altered-context feature itself as the useful normal evidence: with synthetic input or selected positions fixed, altered-context features outperform their clean-view counterparts as both reconstruction targets and memory references. On MVTec memory, the final score change is also spatially selective, with larger reductions on normal patches next to defects than on mid-distance or far-normal patches in all 15 categories. Code is publicly available at \url{https://github.com/jw-chae/boundary_support}.
\end{abstract}

\section{Introduction}
Industrial visual anomaly detection is commonly learned from defect-free images because real defects are sparse, diverse, and expensive to annotate. Memory-based detectors store features from clean training patches and compare test patches with these references \citep{roth2022patchcore,chae2026procon}. Reconstruction-based detectors instead learn to reproduce clean feature targets and use the residual as anomaly evidence \citep{deng2022reverse,guo2025dinomaly}. Strong pretrained encoders improve both families \citep{oquab2024dinov2}, yet the normal evidence they use still comes almost entirely from clean images.

This design hides a simple assumption: \emph{clean training patches are sufficient for the normal regions that appear at test time}. A detector can improve its search rule or reconstruction error only over references and targets that are actually available. If a normal test patch is poorly represented by the clean candidate set, the failure occurs before the final readout.

We therefore ask: \textbf{do clean training images contain all normal patches needed to localize defects?} Most patches in a defective image are still labeled normal, including patches immediately beside the defect. Their local appearance may remain normal while the surrounding image context differs from any clean training image. Because contextual encoders make dense features depend on surrounding content \citep{naseer2021intriguing,darcet2024registers}, these patches can occupy representations that clean-view features do not provide.

We test this with a matched fixed-budget diagnostic. For each donor/evaluation partition, the clean-only memory and ground-truth-assisted oracle are scored on the identical held-out defect images; any image that supplies oracle candidates is excluded from both arms for that partition. The two arms use the same frozen DINOv2 representation, the same test-time scoring rule, and the same number of references. The only change is whether mask-labeled normal patches from real defective images are allowed into the candidate pool. On MVTec AD, the matched comparison raises P-AP from 73.34 to 76.95. Increasing the reference count by 5.3\% on the same candidate union reaches 76.96, only 0.01 point above the fixed-budget result. The fixed-budget memory therefore retains essentially the entire measured gain.

The useful additional patches are concentrated near defects. Restricting the oracle pool to ground-truth-normal patches within two patch cells of the annotated defect reaches 76.76 P-AP, recovering 94.70\% of the fixed-budget gain. Detailed distance sweeps in the two largest-gain categories fall from near to mid to far. The gain also persists after inpainting in those categories and transfers across the tested source/target defect types, reducing the explanatory range of visible-defect leakage and exact defect memorization.

This observation motivates \bs{}. Rather than learning the synthetic core as an anomaly, \bs{} uses it to change the context around a pixel-preserved normal patch. Tokens touched by the nominal insertion or by a directly measured RGB change are excluded. Only neighboring token positions that remain outside this changed set are used as additional normal references or reconstruction targets. In the memory branch, altered-context features compete with clean candidates under the original fixed bank budget. In the reconstruction branch, the same preserved positions receive an auxiliary normal target while the original clean branch is unchanged.

Across MVTec, VisA, and Real-IAD, both branches improve P-AP in all six detector--dataset settings over three paired seeds. The main question then becomes \emph{what part of the intervention is responsible?} Two matched controls answer this directly. For reconstruction, fixing the synthetic input and ring while replacing the target $E(\tilde{x})_p$ with the clean-view feature $E(x)_p$ reverses the gain. For memory, fixing the selected positions and bank budget while replacing $F(\tilde{x})_p$ with $F(x)_p$ lowers P-AP in 26 of 27 categories across MVTec and VisA. A final distance-resolved analysis shows that the learned score change is strongest on normal patches adjacent to real defects.

Our contributions are:
\begin{itemize}
    \item A fixed-budget diagnostic showing that clean-only memory can miss useful normal references: adding mask-labeled normal patches from real defective images raises MVTec P-AP from 73.34 to 76.95, and a two-cell near-defect band recovers 94.70\% of the gain.
    \item \bs{}, a clean-only intervention that changes surrounding context, excludes modified tokens, and uses only preserved neighboring features as memory references or reconstruction targets.
    \item Cross-paradigm and matched-control evidence showing that altered-context features are more useful localization targets/references than their clean-view counterparts, together with a spatial test linking the final score change to normal regions near defects.
\end{itemize}

\section{Related Work}

\paragraph{Normal references and reconstruction targets.}
PatchCore compresses clean patch features into a representative coreset \citep{roth2022patchcore}, while ProCon uses multiple memory anchors for non-parametric feature reconstruction \citep{chae2026procon}. Reconstruction methods instead learn clean feature targets, from reverse distillation \citep{deng2022reverse} to Transformer reconstruction in Dinomaly \citep{guo2025dinomaly}. Across these formulations, normal references or targets are drawn primarily from clean training images.

\paragraph{Learning with contaminated training data.}
SoftPatch filters high-outlier-score patches before coreset construction and reweights retained references \citep{jiang2022softpatch}. InReaCh identifies reliable nominal patches through recurrence across training realizations \citep{mcintosh2023inreach}, while FUN-AD reconstructs a normal memory from potentially contaminated unlabeled data \citep{im2025funad}. SoftPatch+ further strengthens patch-level noise identification with multiple discriminators \citep{wang2025softpatchplus}. These methods suppress unreliable content already present in the training pool; we study normal features that are absent from an otherwise clean pool.

\paragraph{Synthetic anomaly learning.}
CutPaste learns from pasted transformations \citep{li2021cutpaste}; DRAEM reconstructs clean images while segmenting synthetic defects \citep{zavrtanik2021draem}; NSA creates synthetic anomalies by Poisson blending \citep{schluter2022nsa}; and DeSTSeg distills clean teacher features from corrupted inputs \citep{zhang2023destseg}. GLASS and PGBL further synthesize difficult anomalies near normal or prototype boundaries \citep{chen2024glass,liao2026pgbl}. In these methods, the synthetic change serves as anomaly supervision, a localization signal, or a clean-restoration target.

\paragraph{Normal evidence inside defect-containing images.}
MuSc exploits recurrence across unlabeled test images \citep{li2024musc}, while INP-Former and INP-Former++ extract intrinsic normal prototypes from the current image \citep{luo2025inpformer,luo2025inpformerpp}. These methods show that useful normal evidence can remain inside images containing anomalies; our diagnostic instead asks whether such normal evidence is already represented by the clean training pool.

\section{Clean Images Miss Useful Normal Patches Near Defects}
\label{sec:discovery}
We first perform a fixed-budget diagnostic that changes the candidate set while leaving the detector unchanged. Let $\mathcal Z_{\mathrm{clean}}$ be patch features extracted from clean training images and $\mathcal Z_{\mathrm{def},N}$ be patch features labeled normal by the ground-truth mask of real defective images. With bank size $B$,
\begin{align}
M_B^{\mathrm{clean}} &= \KCenter_B(\mathcal Z_{\mathrm{clean}}), \\
M_B^{\mathrm{oracle}} &= \KCenter_B(\mathcal Z_{\mathrm{clean}}\cup\mathcal Z_{\mathrm{def},N}).
\end{align}
The encoder, feature representation, final number of references, and test-time score are fixed. For each donor/evaluation partition, both the clean-only and oracle memories are evaluated on the same held-out images; images supplying oracle candidates are excluded from both arms. Reported values aggregate only these matched held-out evaluations.

\subsection{Changing candidate identity at a fixed memory budget}
On MVTec AD, the clean-only memory obtains 73.34 P-AP. The fixed-budget GT-normal oracle obtains 76.95, a gain of 3.62 points. An expanded oracle that uses the same candidate union but 5.3\% more stored references reaches 76.96. The expanded and fixed-budget results differ by 0.01 point, while image AUROC changes from 99.76 to 99.71. At the tested budget, the main change is therefore the identity of the available references, and the resulting effect is concentrated in pixel localization.

\subsection{Most of the gain comes from patches near defects}
For a ground-truth-normal patch $p$, let $\mathcal A_{\mathrm{gt}}$ denote the annotated defect cells and let $d(p,\mathcal A_{\mathrm{gt}})$ be the Euclidean patch-grid distance to the nearest cell in $\mathcal A_{\mathrm{gt}}$. Restricting the additional candidates to $0<d\le2$ reaches 76.76 P-AP over all 15 categories and recovers 94.70\% of the fixed-budget gain. In the two categories with the largest oracle gains, the same fixed-budget comparison gives $+16.29/+6.25/+0.55$ points for near/mid/far patches in leather and $+13.33/+2.14/+0.67$ in tile. Thus the all-category statement is the near-band recovery; the full near--mid--far ordering is a two-category diagnostic.

\begin{figure}[t]
  \centering
  \includegraphics[width=\textwidth]{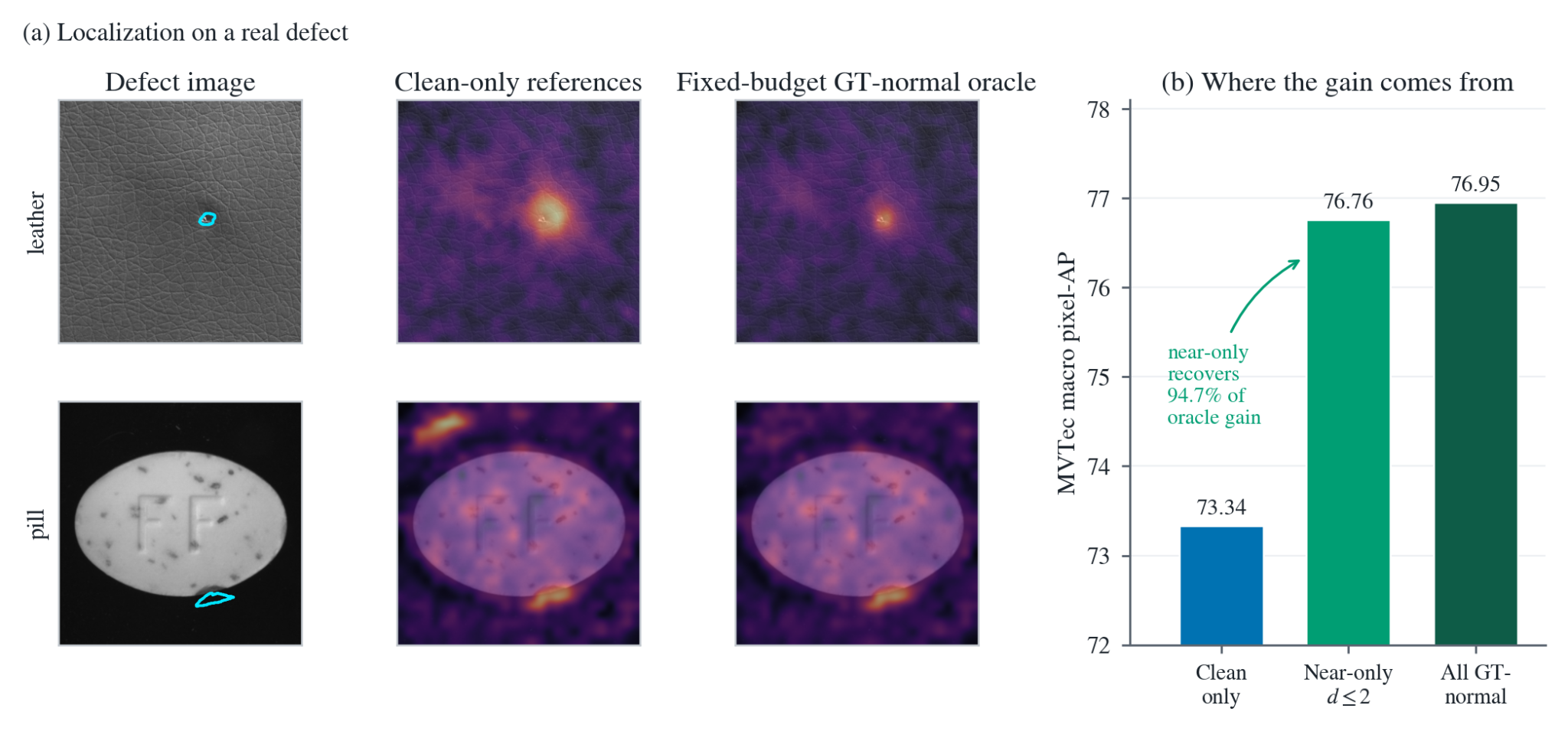}
  \caption{\textbf{Clean images miss useful normal patches near defects.} Left: two real-defect examples under the clean-only memory and the fixed-budget GT-normal oracle. Right: MVTec macro P-AP at a fixed reference count. Ground-truth-normal candidates within two patch cells recover 94.70\% of the oracle gain. The oracle uses ground-truth masks only as a diagnostic.}
  \label{fig:discovery}
\end{figure}

\subsection{Visible defect signal and exact defect identity are insufficient explanations}
We test two simple alternatives in leather and tile. First, we dilate and inpaint the annotated defect before re-encoding the image. The resulting candidate features still improve over the clean memory by 11.72 and 6.11 P-AP points, respectively. Second, ground-truth-normal candidate features collected beside one defect type are evaluated only on other defect types. Three tested source-to-target directions remain positive: $+7.69$ points for leather cut $\rightarrow$ other types, $+0.79$ for tile crack $\rightarrow$ other types, and $+1.46$ for tile glue-strip $\rightarrow$ other types. These scoped controls narrow the explanation: the remaining gains after inpainting and cross-defect transfer make visible-defect signal and exact same-defect memorization insufficient on their own.

\subsection{A small near-defect pool receives disproportionate memory capacity}
The near-defect cohort is numerically small, yet approximate $k$-center selection repeatedly promotes it. In leather, near-defect GT-normal patches form only 0.23\% of the oracle candidate pool but 2.68\% of the selected anchors, an 11.73$\times$ enrichment. In tile, the corresponding shares are 0.70\% and 3.44\%, a 4.95$\times$ enrichment. The fixed-budget oracle improves 14 of 15 categories, and the two-cell near-only oracle improves 13 of 15. These observations explain how a small candidate cohort can materially change a fixed-size bank: the selection geometry allocates it more memory than its raw frequency would suggest. The downstream P-AP comparison, rather than enrichment alone, establishes that those replacements are useful.

\section{BoundarySupport: Learning Preserved Patches under Altered Context}
\label{sec:method}
Section~\ref{sec:discovery} identifies a useful population but uses real defective images and masks. \bs{} asks whether clean training images can expose useful normal features by changing only their surrounding context.

\subsection{Exclude modified tokens; learn from preserved neighbors}
For a clean image $x\in\{0,\ldots,255\}^{H\times W\times3}$, a procedural intervention $g$ produces a synthetic view and nominal insertion mask,
\begin{equation}
(\tilde{x},A)=g(x).
\end{equation}
We use CutPaste-Scar \citep{li2021cutpaste}, same-category Poisson paste following NSA \citep{schluter2022nsa}, and a same-image local warp followed by Poisson paste. All three interventions are fixed procedural transformations.

Because interpolation and blending can modify pixels outside $A$, we compare the two uint8 RGB images directly:
\begin{equation}
D(u)=\mathbf 1\!\left[\frac{1}{3}\sum_{c=1}^{3}\left|\tilde{x}_c(u)-x_c(u)\right|>1\right].
\label{eq:change}
\end{equation}
Let $P$ max-pool a pixel mask to the encoder token grid. The excluded token set and retained ring are
\begin{align}
C &= P(A)\lor P(D), \\
R &= \operatorname{Dilate}(P(A),r)\setminus C, \qquad r=2.
\label{eq:ring}
\end{align}
Dilation is applied to $P(A)$ rather than to $C$. The complete synthetic image is still encoded, so the synthetic insertion remains visible as context, while no token in $C$ can become a normal memory candidate or direct reconstruction target. Only positions in $R$ are used as additional normal memory candidates or reconstruction targets. These positions are pixel-preserved, while their encoder features are computed from the complete altered-context view.

The detected-change term $P(D)$ is operationally important because blending and interpolation can touch token footprints outside the nominal insertion. In the seed-0 reconstruction control, the strict construction above improves P-AP by 4.94 points on MVTec and 2.09 on VisA. Using the same synthetic view while excluding only the nominal mask yields $+0.70$ on MVTec and $-2.40$ on VisA. Thus the method does not treat an arbitrary geometric ring as preserved: it removes token footprints that the generated image actually changed according to Eq.~\ref{eq:change}.

\begin{figure}[t]
  \centering
  \includegraphics[width=\textwidth]{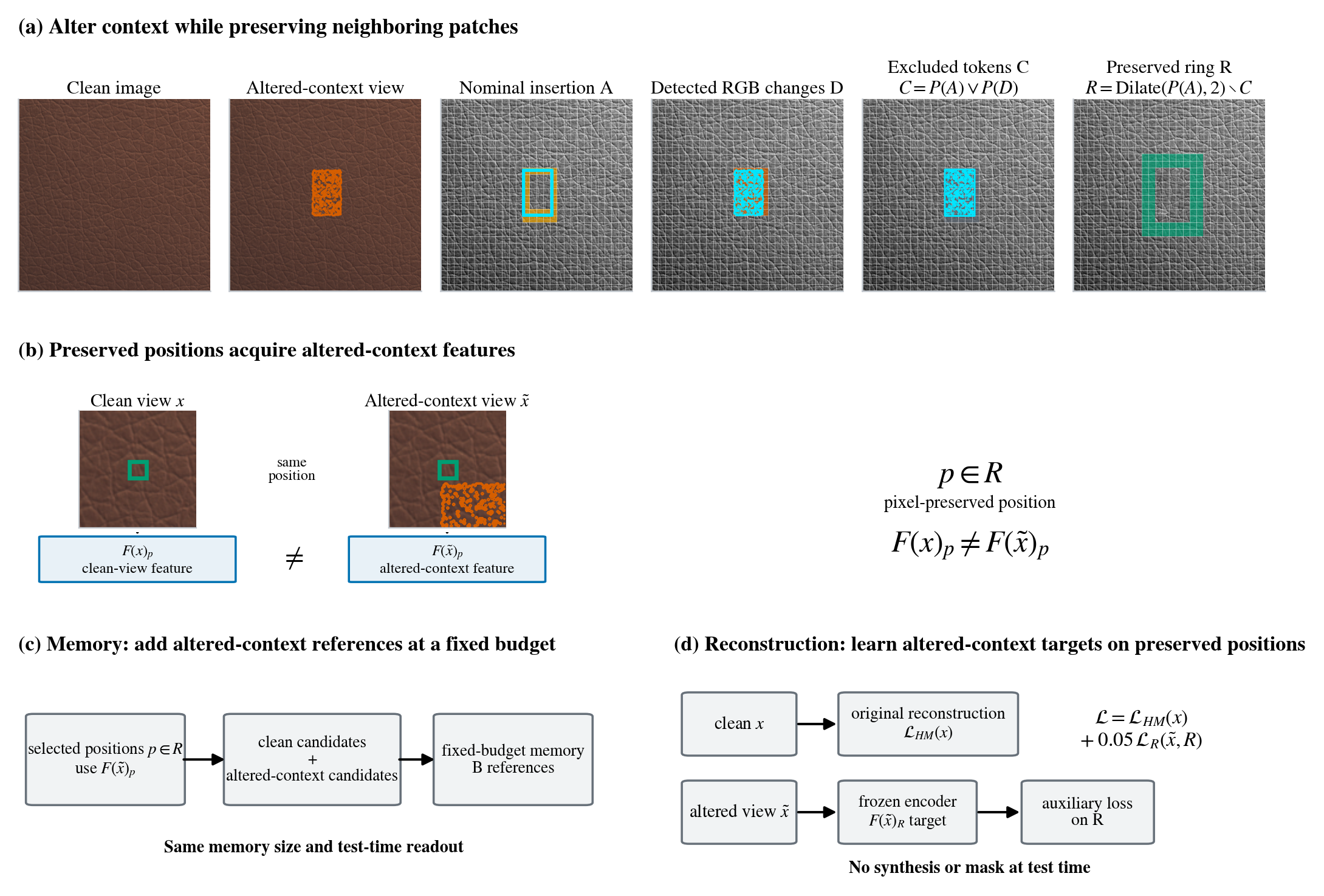}
  \caption{\textbf{BoundarySupport exposes normal features under altered context while learning only from pixel-preserved positions.} \textbf{(a)} A procedural intervention produces an altered-context view $\tilde{x}$ and nominal insertion mask $A$. Tokens intersecting the nominal insertion or detected RGB changes form the excluded set $C$, while the surrounding pixel-preserved positions form the ring $R$. \textbf{(b)} At a pixel-preserved position $p\in R$, the clean view and altered-context view yield the clean-view feature $F(x)_p$ and altered-context feature $F(\tilde{x})_p$, respectively. \textbf{(c)} The memory branch adds selected altered-context features to the clean candidate pool and selects exactly $B$ references, preserving the original memory budget and test-time readout. \textbf{(d)} The reconstruction branch keeps the original clean objective and adds altered-context encoder targets only at positions in $R$. Test-time scoring uses the observed image without synthesis or masks.}
  \label{fig:method}
\end{figure}

\subsection{Memory branch}
The memory branch uses frozen DINOv2 features and the ProCon test-time readout \citep{chae2026procon}. Candidate discovery uses a separate score. For layer $\ell$, view $v$, and matched position $p$,
\begin{align}
d_\ell(v,p) &= \operatorname{median}_{b=1,\ldots,5}\min_{m\in M_{\ell,b}}\|F_\ell(v)_p-m\|_2, \\
s_{\mathrm{cand}}(v,p) &= \frac{1}{|\mathcal L|}\sum_{\ell\in\mathcal L}d_\ell(v,p), \\
\Delta_p &= s_{\mathrm{cand}}(\tilde{x},p)-s_{\mathrm{cand}}(x,p).
\end{align}
Candidate discovery uses $k=1$ Euclidean nearest distance in each bank; the deployed ProCon readout remains $k=5$. The two filters have distinct roles. The clean-view condition keeps positions already compatible with the clean memory, while the $\Delta_p$ condition identifies positions whose representation moves away from that memory after context is altered. Quantiles are computed within each intervention type so that one intervention family cannot dominate through scale alone. The uniform policy used in the main results retains
\begin{equation}
p\in R,\quad s_{\mathrm{cand}}(x,p)\le Q_{0.50},\quad \Delta_p\ge Q_{0.90},\quad \Delta_p>0.
\label{eq:memselect}
\end{equation}
A shared location mask selects the altered-context feature at every layer. If $\widetilde{\mathcal Z}^{\mathrm{sel}}_\ell$ denotes the selected altered-context features and $\mathcal Z^{\mathrm{clean}}_\ell$ the clean candidates, each bank is rebuilt as
\begin{equation}
M'_{\ell,b}=\KCenter_B^{(b)}\!\left(\mathcal Z^{\mathrm{clean}}_\ell\cup\widetilde{\mathcal Z}^{\mathrm{sel}}_\ell\right).
\end{equation}
The target size $B$ is exactly the clean-baseline bank size.

\subsection{Reconstruction branch}
The reconstruction branch follows the frozen-encoder Dinomaly framework \citep{guo2025dinomaly}. The encoder provides fixed feature targets; only the bottleneck and decoder are trained. The original clean hard-mined objective $\mathcal L_{\mathrm{HM}}(x)$ is unchanged. For the two fused encoder targets $E_j$ and decoder outputs $G_j$, the synthetic branch adds
\begin{equation}
\mathcal L_R=\frac{1}{2}\sum_{j=1}^{2}\operatorname{mean}_{p\in R}\left[1-\cos\!\left(\sg(E_j(\tilde{x})_p),G_j(\tilde{x})_p\right)\right],
\end{equation}
with total objective
\begin{equation}
\mathcal L=\mathcal L_{\mathrm{HM}}(x)+0.05\,\mathcal L_R.
\end{equation}
One of the three interventions is sampled uniformly for each training example. Positions in $C$ receive no direct loss but remain visible to contextual attention. Test-time scoring is the original reconstruction residual on the observed image.

\section{Experiments}
\label{sec:experiments}
\subsection{Setup}
We evaluate on MVTec AD (15 categories) \citep{bergmann2019mvtec}, VisA (12) \citep{zou2022visa}, and Real-IAD (30) \citep{wang2024realiad}. Memory banks use 5\% of clean training patches on MVTec/VisA and 1\% on Real-IAD; each \bs{} bank has exactly the corresponding clean-baseline count. Reconstruction trains for 5,000 iterations per category. Real-IAD synthesis donors are restricted to the same category and camera view. All six detector--dataset settings use paired seeds 0/1/2. Hyperparameters were developed on MVTec and VisA; the Real-IAD configuration was fixed before its outcomes were inspected.

P-AP from the full pixel ranking is our primary endpoint. Table~\ref{tab:main} also reports I-AUROC, P-AUROC, P-F1, and AUPRO. The appendix provides seed-0 values for all seven metrics, three-seed P-AP summaries, and category results. Values are 0--100 category macros; uncertainty is over three paired seed macros, and category--seed counts report sign consistency.

\subsection{P-AP improves in all six detector--dataset settings}
\begin{table}[t]
\centering
\small
\setlength{\tabcolsep}{2.8pt}
\caption{Paired changes after adding \bs{} (mean$\pm$sample standard deviation over three seed-level category macros).}
\label{tab:main}
\begin{tabular}{llrrrrr}
\toprule
Branch & Dataset & $\Delta$ I-AUROC & $\Delta$ P-AUROC & $\Delta$ P-AP & $\Delta$ P-F1 & $\Delta$ AUPRO \\
\midrule
Memory & MVTec & $-0.03\pm0.02$ & $+0.07\pm0.01$ & $\mathbf{+2.21\pm0.04}$ & $+1.31\pm0.01$ & $+0.22\pm0.01$ \\
Memory & VisA & $+0.02\pm0.01$ & $-0.02\pm0.01$ & $\mathbf{+1.82\pm0.03}$ & $+1.03\pm0.02$ & $-0.03\pm0.01$ \\
Memory & Real-IAD & $-0.11\pm0.02$ & $-0.01\pm0.01$ & $\mathbf{+1.76\pm0.04}$ & $+1.31\pm0.02$ & $-0.05\pm0.01$ \\
Recon. & MVTec & $-0.08\pm0.03$ & $+0.16\pm0.02$ & $\mathbf{+5.01\pm0.07}$ & $+2.97\pm0.14$ & $+0.55\pm0.04$ \\
Recon. & VisA & $+0.05\pm0.01$ & $-0.16\pm0.03$ & $\mathbf{+2.27\pm0.38}$ & $+1.16\pm0.28$ & $+0.21\pm0.18$ \\
Recon. & Real-IAD & $-0.52\pm0.04$ & $-0.03\pm0.01$ & $\mathbf{+4.20\pm0.12}$ & $+2.56\pm0.09$ & $+0.16\pm0.03$ \\
\bottomrule
\end{tabular}
\end{table}

Table~\ref{tab:main} shows the same localization direction in all six settings. Memory improves all 171 category--seed pairs (45/45 MVTec, 36/36 VisA, 90/90 Real-IAD); reconstruction improves 158/171 (45/45, 29/36, 84/90). Overall, P-AP rises in 329/342 pairs. Real-IAD reconstruction gains $4.20\pm0.12$ P-AP while I-AUROC falls $0.52\pm0.04$, so the measured effect is a pixel-localization ranking improvement.

\subsection{Altered-context features are the effective normal evidence}
\label{sec:controls}
We next isolate feature identity with two matched controls. In reconstruction, the synthetic input $\tilde{x}$, ring $R$, frozen encoder, trainable network, auxiliary weight, training length, and seed are fixed. Only the ring target changes from $\sg(E(\tilde{x})_p)$ to the clean-view feature $\sg(E(x)_p)$. In memory, the selected locations $\mathcal P_{\mathrm{sel}}$, clean candidate pool, absolute bank budget, joint $k$-center selection, projections, inference, and seed are fixed. Only the selected feature changes from $F(x)_p$ to $F(\tilde{x})_p$.

\begin{table}[t]
\centering
\small
\setlength{\tabcolsep}{3.7pt}
\caption{Matched feature-identity controls at seed 0. Reconstruction fixes the synthetic input and ring while changing only the reconstruction target; memory fixes the selected positions and bank construction while changing only the stored feature.}
\label{tab:mechanism}
\begin{tabular}{llrrrr}
\toprule
Branch & Dataset & Clean-view target/reference & Altered-context target/reference & Altered $-$ clean & Wins \\
\midrule
Reconstruction & MVTec & 61.82 & \textbf{73.89} & \textbf{+12.07} & 15/15 \\
Reconstruction & VisA  & 44.34 & \textbf{52.79} & \textbf{+8.45}  & 11/12 \\
Memory & MVTec & 73.36 & \textbf{75.60} & \textbf{+2.23} & 15/15 \\
Memory & VisA  & 52.28 & \textbf{54.11} & \textbf{+1.84} & 11/12 \\
\bottomrule
\end{tabular}
\end{table}

The reconstruction control is especially discriminative. Restoring the clean-view feature from the same synthetic input lowers P-AP by 7.13 points on MVTec and 6.36 on VisA relative to baseline; using the feature actually formed under altered context instead raises P-AP by 4.94 and 2.09. The direct altered-versus-clean target gap is 12.07 and 8.45 points. This matched comparison separates target identity from synthetic exposure.

The memory control separates spatial mining from feature identity. At exactly the same selected locations, altered-context features outperform the clean-view features by 2.23 points on MVTec and 1.84 on VisA, with positive category-level differences in 26 of 27 categories. AUPRO and image AUROC remain mixed or nearly unchanged in this comparison, while the P-AP localization difference is consistent.

\subsection{Where does the final score change?}
The discovery in Section~\ref{sec:discovery} is spatial: useful oracle references concentrate near defects. We therefore ask whether the final detector changes the same region. For the MVTec memory branch, each ground-truth-normal test patch is assigned its distance $d$ from the nearest defect patch, and we compute
\begin{equation}
\Delta s(p)=s_{\bs}(p)-s_{\mathrm{baseline}}(p).
\end{equation}
Scores are reported below on a $100\Delta s$ scale.

\begin{table}[t]
\centering
\small
\caption{Distance-resolved MVTec memory score changes at seed 0.}
\label{tab:distance_score}
\begin{tabular}{lr}
\toprule
Spatial cohort & Mean $100\Delta s$ \\
\midrule
Defect patches & $-0.61$ \\
Near-defect normal ($0<d\le2$) & $-0.37$ \\
Mid-distance normal ($2<d\le4$) & $+0.11$ \\
Far normal ($d>4$) & $+0.19$ \\
Clean-image normal & $+0.21$ \\
\bottomrule
\end{tabular}
\end{table}

The score change is spatially selective. Defect and near-defect normal patches decrease on average, whereas mid-distance, far-normal, and clean-image normal patches increase slightly. More importantly, the within-category near-minus-far and near-minus-mid contrasts are negative in all 15 categories; their median $100\Delta s$ contrasts are $-0.51$ and $-0.43$, respectively. The final memory therefore changes normal responses preferentially near defects, the same spatial region identified by the fixed-budget oracle. Because far-normal and clean-image normal scores move slightly upward rather than downward, this pattern is not a uniform suppression of normal scores.

\begin{figure}[t]
  \centering
  \includegraphics[width=\textwidth]{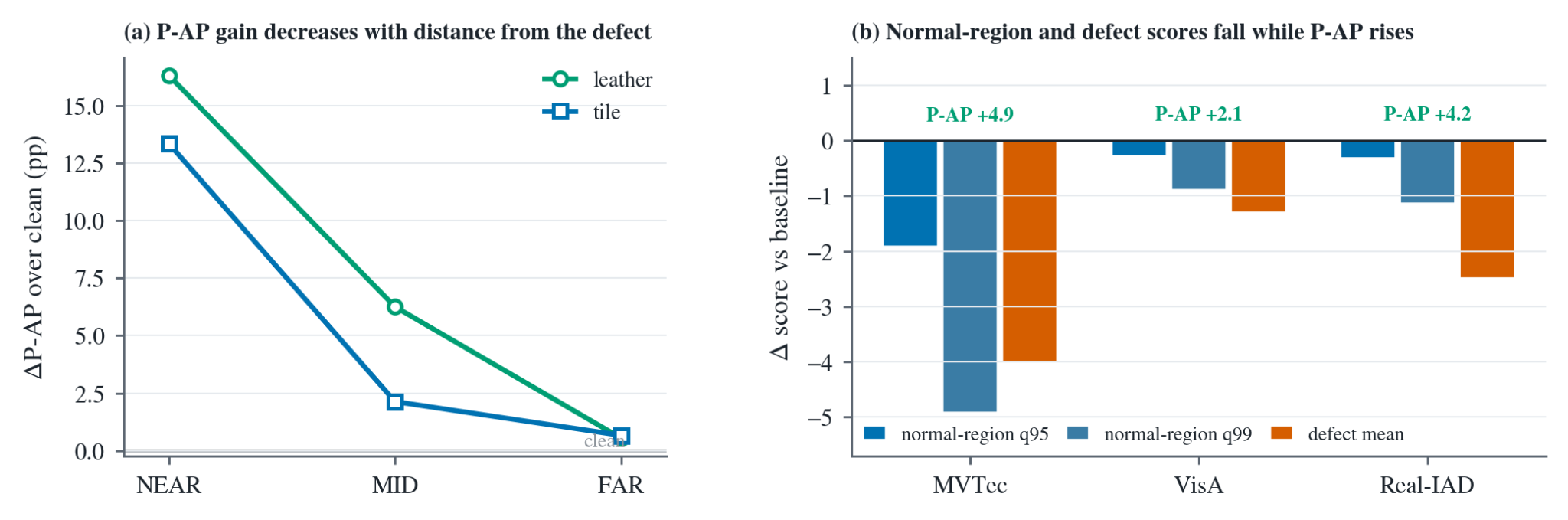}
  \caption{\textbf{Complementary spatial and score diagnostics.} Left: in leather and tile, the fixed-budget oracle gain decreases from near to mid to far candidate bands. Right: reconstruction at seed 0 reduces high anomaly scores on normal regions and also reduces mean defect scores while P-AP rises. These summaries diagnose where scores move; P-AP is determined by the full ranking.}
  \label{fig:diag}
\end{figure}

A complementary reconstruction diagnostic is shown in Figure~\ref{fig:diag}. At seed 0, normal-region q95/q99 change by $-1.91/-4.91$ on MVTec, $-0.28/-0.88$ on VisA, and $-0.32/-1.13$ on Real-IAD (0--100 scale). Mean defect scores also decrease by 4.01, 1.29, and 2.49 points, while P-AP rises by 4.94, 2.09, and 4.20. The full three-seed MVTec and VisA reconstruction runs repeat the joint decrease of these recorded score summaries. Localization therefore does not improve by simply increasing defect activation; it improves when the full ordering of defect and normal responses becomes more favorable.

Together, the oracle and matched controls identify the same object across training and inference: pixel-preserved normal patches acquire useful altered-context features, and the resulting detector changes normal responses most strongly near real defects.

\subsection{Failure cases expose the same ranking trade-off}
The aggregate localization gain does not require every category--seed pair to improve. Across three paired seeds, memory improves all 171 category--seed pairs, whereas reconstruction improves 158 of 171. At seed 0, reconstruction improves 52 of 57 categories; the remaining cases are informative because they occur in the same score direction as the successful cases. On VisA, cashew loses 2.79 P-AP points while its recorded normal-region q99 falls by 0.74 and its mean defect score falls by 1.79. PCB3 and fryum lose 2.42 and 1.26 points, respectively. On Real-IAD, rolled strip base loses 2.92 P-AP points even though its pixel AUROC and AUPRO change only slightly.

These cases show why the method is evaluated by the complete pixel ranking rather than by one score summary. Lower scores on difficult normal regions are useful only when defect evidence is preserved enough to improve their ordering against defect pixels. This also explains the Real-IAD reconstruction result in Table~\ref{tab:main}: a 4.20-point P-AP gain coexists with a 0.52-point decrease in image AUROC. BoundarySupport changes which normal evidence is learned; it does not impose a monotone improvement on every metric or category.

\section{Discussion}
\paragraph{Reference and target identity are part of the detector.}
The fixed-budget oracle changes candidate identity while holding the representation, test-time score, and reference count fixed. The 3.62-point P-AP gain makes the composition of normal evidence a first-order part of the detector, alongside the readout applied after a bank or reconstruction target set is defined.
Contaminated-training methods address the opposite failure mode by removing unreliable patches from a candidate pool. Our results show that even a clean pool can omit useful normal features. Memory construction therefore depends on both which candidates are excluded and which normal representations are available.

\paragraph{Normal-pool completeness is representation-dependent.}
Pixel-preserved patches can acquire different representations when their surrounding context changes, so image-level cleanliness does not imply feature-level coverage. For contextual encoders, normal evidence depends on which contextual states of normal patches are represented, not only on which source images are nominal.

\paragraph{The real-defect diagnostic and the clean-only intervention play different roles.}
The oracle uses real defects only to locate a deficit in clean references. BoundarySupport instead starts from a known-clean image, creates a controlled context change, measures which pixel footprints changed, and excludes the corresponding tokens before adding normal references or reconstruction targets. The oracle claim is therefore annotation-level, while the method neither reuses real defect pixels nor relies on physical purity at a real mask boundary. Its matched controls test the representational consequence directly: a pixel-preserved patch becomes more useful when encoded under altered context.

\paragraph{The same feature matters in two different computations.}
The matched controls narrow this statement further. Reconstruction performs better when it learns the preserved feature encoded under altered context rather than the matched clean-view feature. Memory shows the same ordering when location selection is fixed. These two computations use the feature differently---one as a training target, one as a stored reference---yet both identify the altered-context representation as the more useful localization evidence in the tested settings.

\paragraph{Localization gain and defect evidence can trade off.}
\bs{} improves P-AP without uniformly improving image-level metrics. Reconstruction can reduce true-defect scores together with high normal-region scores, and negative categories appear when defect suppression dominates. A natural next step is therefore to preserve defect ordering explicitly while exposing missing normal evidence. Another open problem is to identify deployment-relevant missing patches directly, without relying on a synthetic context intervention.

The current evidence covers DINOv2 ViT-B-family representations in one memory branch and one reconstruction branch. The two-paradigm agreement establishes transfer across different normal-modeling computations in this representation family; other encoders remain an empirical question.

\section{Conclusion}
At a fixed memory budget, ground-truth-normal patches from real defective images provide useful references absent from a clean-only pool: MVTec P-AP rises from 73.34 to 76.95, and patches within two cells recover 94.70\% of the gain. \bs{} brings this observation to clean-only training by altering context, excluding modified tokens, and learning from preserved neighbors.

Across memory and reconstruction on three datasets and three paired seeds, \bs{} consistently improves pixel localization. Matched controls show that altered-context features outperform their clean-view counterparts as both targets and references. Normal-only detection therefore depends on which normal features are available, not only on how they are scored.

\bibliographystyle{iclr2027_conference}
\bibliography{references}

\clearpage
\appendix
\section{Experimental Setup}
\label{app:setup}

\subsection{Datasets and evaluation}
We use the standard normal-only training splits and labeled test splits of MVTec AD \citep{bergmann2019mvtec}, VisA \citep{zou2022visa}, and Real-IAD \citep{wang2024realiad}. MVTec and VisA were used to configure the method. For Real-IAD, the generator family, ring radius, reconstruction weight, and memory ratio were fixed before inspecting evaluation outcomes. All six detector--dataset settings were then evaluated with seeds 0, 1, and 2. Real-IAD synthesis donors are restricted to the same category and camera view.

\begin{table}[h]
\centering
\small
\caption{Evaluation settings.}
\label{tab:app_setup}
\begin{tabular}{lrrrr}
\toprule
Dataset & Categories & Memory ratio & Memory seeds & Reconstruction seeds \\
\midrule
MVTec AD & 15 & 5\% & 0,1,2 & 0,1,2 \\
VisA & 12 & 5\% & 0,1,2 & 0,1,2 \\
Real-IAD & 30 & 1\% & 0,1,2 & 0,1,2 \\
\bottomrule
\end{tabular}
\end{table}

The VisA memory result uses one uniform rule for all 12 categories and all three seeds: $q_{\mathrm{clean}}=0.50$, $q_{\Delta}=0.90$, all three interventions, joint candidate selection, and the original fixed 5\% bank. Every layer and bank has the same target size as the clean baseline. The MVTec GT-assisted diagnostic uses seed 0. For each donor/evaluation partition, the clean-only and oracle variants are evaluated on the identical held-out images; donor images are excluded from both arms for that partition. Reported oracle diagnostics aggregate only these matched held-out evaluations.

\subsection{Input geometry and scoring}
All encoders receive ImageNet-normalized RGB crops. For memory, MVTec images are aspect-preservingly resized to 448 and center-cropped to $392\times392$; VisA and Real-IAD use a square 448 resize followed by the same crop. Reconstruction uses the square resize/crop on all three datasets. Patch size 14 gives a $28\times28$ token grid.

The memory branch averages four layer maps, resizes to 392, applies Gaussian smoothing with $\sigma=4$, and uses the mean of the highest 0.5\% patch scores as the image score. The reconstruction branch averages two fused residual maps, bilinearly resizes to $256\times256$, applies a $5\times5$ Gaussian filter with $\sigma=4$, and uses the mean of the highest 1\% pixel scores as the image score. BoundarySupport does not change the test-time input or scoring path.

\subsection{Metrics and aggregation}
All metrics are reported on a 0--100 scale. Image metrics are I-AUROC, I-AP, and maximum I-F1 along the image precision--recall curve. Pixel metrics are P-AUROC, P-AP, and maximum P-F1 after flattening evaluated pixels within each category. AUPRO integrates connected-component overlap against false-positive rate up to FPR 0.30 using 200 thresholds. Dataset results are arithmetic means over categories. Multi-seed entries first compute the category macro for each seed and then report the paired mean and sample standard deviation ($\mathrm{ddof}=1$). Category--seed counts summarize sign consistency and are not treated as independent replicates for uncertainty estimation.

\section{Fixed-Budget GT-Normal Diagnostic}
\label{app:oracle}
The diagnostic uses a frozen DINOv2-B/14 representation, the five-bank memory readout, a 5\% memory ratio, and the evaluation geometry above. Ground-truth masks identify normal-labeled patches inside real defective images. Clean and oracle variants are always compared on the same held-out images within each donor/evaluation partition.

\subsection{Candidate identity and memory budget}
The clean-only memory builds its candidate pool only from clean training patches. The expanded GT-normal oracle adds normal-labeled patches from defective images and applies the same percentage memory ratio, increasing the final bank size by about 5\%. The fixed-budget GT-normal oracle uses the same enlarged candidate pool but restores the category-specific absolute bank count of the clean-only memory.

\begin{table}[h]
\centering
\small
\caption{MVTec fixed-budget diagnostic on matched held-out evaluation sets.}
\label{tab:app_oracle_macro}
\begin{tabular}{lrrrrr}
\toprule
Candidate pool & I-AUROC & P-AUROC & P-AP & AUPRO & Rel. bank size \\
\midrule
Clean candidates (matched eval.) & 99.76 & 98.80 & 73.34 & 96.18 & 1.00 \\
Clean + GT-normal, expanded & 99.71 & 98.94 & 76.96 & 96.64 & 1.05 \\
Clean + GT-normal, fixed budget & 99.71 & 98.94 & 76.95 & 96.64 & 1.00 \\
\bottomrule
\end{tabular}
\end{table}

The fixed-budget candidate change raises P-AP by 3.62 points. Expanded and fixed-budget banks differ by only 0.01 point P-AP at this budget.

\begin{table}[!htbp]
\centering
\scriptsize
\caption{MVTec category P-AP for the clean-only memory and GT-normal oracle variants.}
\label{tab:app_oracle_category}
\begin{tabular}{lrrrr}
\toprule
Category & Clean (matched eval.) & Expanded GT-normal & Fixed-budget GT-normal & Near-defect GT-normal \\
\midrule
bottle & 89.02 & 90.29 & 90.30 & 90.21 \\
cable & 76.70 & 77.36 & 77.32 & 77.34 \\
capsule & 64.40 & 63.87 & 63.88 & 63.47 \\
carpet & 80.07 & 84.04 & 84.05 & 83.99 \\
grid & 63.74 & 66.38 & 66.45 & 66.26 \\
hazelnut & 81.64 & 84.70 & 84.62 & 84.67 \\
leather & 61.97 & 77.52 & 77.55 & 78.26 \\
metal nut & 83.07 & 87.38 & 87.36 & 87.09 \\
pill & 75.61 & 76.98 & 76.97 & 76.31 \\
screw & 65.89 & 67.00 & 67.01 & 66.12 \\
tile & 76.69 & 89.75 & 89.76 & 90.02 \\
toothbrush & 63.00 & 63.41 & 63.38 & 62.77 \\
transistor & 66.84 & 67.65 & 67.58 & 67.26 \\
wood & 79.57 & 82.76 & 82.78 & 83.08 \\
zipper & 71.83 & 75.28 & 75.26 & 74.58 \\
\midrule
Macro & 73.34 & 76.96 & 76.95 & 76.76 \\
\bottomrule
\end{tabular}
\end{table}

The fixed-budget oracle improves 14 of 15 categories. The near-defect variant improves 13 of 15. The clean column here is the matched clean control evaluated under the oracle donor/evaluation partitions; Section~\ref{app:categories} separately reports the ordinary seed-0 full-test category results.

\subsection{Distance from the annotated defect}
For a ground-truth-normal patch $p$ and annotated defect-cell set $\mathcal A_{\mathrm{gt}}$, define
\begin{align}
\mathcal H_{\mathrm{near}} &= \{p:y_p=N,\ 0<d(p,\mathcal A_{\mathrm{gt}})\le2\},\\
\mathcal H_{\mathrm{mid}} &= \{p:y_p=N,\ 2<d(p,\mathcal A_{\mathrm{gt}})\le4\},\\
\mathcal H_{\mathrm{far}} &= \{p:y_p=N,\ d(p,\mathcal A_{\mathrm{gt}})>4\}.
\end{align}
The all-category near-defect oracle obtains 99.73 I-AUROC, 98.92 P-AUROC, 76.76 P-AP, and 96.57 AUPRO. It is 0.19 P-AP below the fixed-budget all-GT-normal oracle and recovers 94.70\% of its gain over the clean-only memory.

\begin{table}[h]
\centering
\small
\caption{Distance-stratified fixed-budget P-AP in the two largest-gain categories.}
\label{tab:app_distance_oracle}
\begin{tabular}{lrrrrr}
\toprule
Category & Clean & Fixed-budget GT-normal & Near & Mid & Far \\
\midrule
leather & 61.97 & 77.55 & 78.26 & 68.22 & 62.52 \\
tile & 76.69 & 89.76 & 90.02 & 78.83 & 77.36 \\
\bottomrule
\end{tabular}
\end{table}

\subsection{Inpainting and cross-defect transfer}
We dilate the annotated defect by seven pixels, apply Telea inpainting with radius 3, re-encode the image, and retain candidate patches labeled normal by the original mask. The final memory count remains fixed.

\begin{table}[h]
\centering
\small
\caption{P-AP after removing the visible annotated defect by inpainting.}
\begin{tabular}{lrrr}
\toprule
Category & Clean & Fixed-budget GT-normal & Inpainted-defect GT-normal \\
\midrule
leather & 61.97 & 77.55 & 73.69 \\
tile & 76.69 & 89.76 & 82.80 \\
\bottomrule
\end{tabular}
\end{table}

For cross-defect transfer, GT-normal candidates are collected next to one source defect type and evaluation is restricted to other defect types.

\begin{table}[h]
\centering
\small
\caption{Cross-defect transfer P-AP.}
\begin{tabular}{lrrr}
\toprule
Source $\rightarrow$ evaluation defects & Clean & Cross-defect GT-normal & $\Delta$P-AP \\
\midrule
leather: cut $\rightarrow$ color/fold/glue/poke & 64.65 & 72.35 & +7.69 \\
tile: crack $\rightarrow$ other types & 90.65 & 91.44 & +0.79 \\
tile: glue-strip $\rightarrow$ other types & 73.98 & 75.44 & +1.46 \\
\bottomrule
\end{tabular}
\end{table}

\subsection{Reference selection and oracle score changes}
Near-defect patches occupy a small fraction of the candidate pool but a larger fraction of selected references.

\begin{table}[h]
\centering
\small
\caption{Near-defect candidate and selected-reference shares.}
\begin{tabular}{lrrrr}
\toprule
Category & Candidate near & Selected near & Enrichment & Selected anchors \\
\midrule
leather & 0.23\% & 2.68\% & 11.73$\times$ & 258 / 9,604 \\
tile & 0.70\% & 3.44\% & 4.95$\times$ & 310 / 9,016 \\
\bottomrule
\end{tabular}
\end{table}

The all-GT-normal fixed-budget oracle similarly raises the selected share of defective-image normal candidates from 4.95\% to 15.98\% in leather and from 4.20\% to 8.65\% in tile.

\begin{table}[h]
\centering
\small
\caption{Raw oracle score summaries in leather and tile.}
\resizebox{\textwidth}{!}{%
\begin{tabular}{llrrrr}
\toprule
Category & Memory & Good normal q95 & Defect-image normal q95 & q99 & Defect mean \\
\midrule
leather & Clean & 0.40 & 0.48 & 0.61 & 0.65 \\
leather & Fixed-budget GT-normal & 0.40 & 0.44 & 0.53 & 0.60 \\
tile & Clean & 0.45 & 0.55 & 0.66 & 0.62 \\
tile & Fixed-budget GT-normal & 0.45 & 0.51 & 0.60 & 0.61 \\
\bottomrule
\end{tabular}}
\end{table}

\section{Synthetic Context Interventions}
\label{app:generators}
The three interventions are fixed, non-learned image operations. Each receives a clean uint8 RGB image $x$, a seeded random number generator, and, when needed, a clean donor $x'$ from the same category. Real-IAD donors are also restricted to the same camera view. Each generator outputs a synthetic view $\tilde{x}$ and nominal insertion mask $A$.

\subsection{CutPaste-Scar}
We use the scar geometry of CutPaste \citep{li2021cutpaste}. Let $s=\min(H,W)/224$. The short side is sampled as an integer from
\[
[\max(2,\operatorname{round}(2s)),\ \max(3,\operatorname{round}(16s)))
\]
and the long side from
\[
[\max(\mathrm{short}+1,\operatorname{round}(10s)),\ \max(\mathrm{short}+2,\operatorname{round}(25s))).
\]
Orientation is swapped with probability 0.5. A same-image patch receives independent channel gains from $U[0.9,1.1]$, rotation from $U[-45^\circ,45^\circ]$, bilinear interpolation, and reflected borders before being pasted at a valid location. The rotated binary support defines $A$.

\subsection{Cross-image Poisson paste}
For the NSA-style intervention \citep{schluter2022nsa}, donor half-sizes are sampled independently as
\[
h_{1/2}=\operatorname{clip}[(0.03+G)H,0.05H,0.20H],\qquad G\sim\operatorname{Gamma}(2,0.05),
\]
with the analogous rule for width. The donor crop is resized by $\operatorname{clip}(Z,0.7,1.3)$ for $Z\sim\mathcal N(1,0.5^2)$ and pasted with OpenCV \texttt{NORMAL\_CLONE}. The clone support excluding its one-pixel border defines $A$.

\subsection{Local warp and Poisson paste}
A same-image crop spanning 10--25\% of image height and width is transformed by rotation $U[-15^\circ,15^\circ]$ and isotropic scale $U[0.82,1.18]$ with bilinear interpolation and reflected borders, then inserted with the same Poisson paste and mask construction.

\subsection{Generation schedule}
Memory materializes clean/synthetic feature pairs; reconstruction generates a synthetic view online. MVTec and Real-IAD memory use one cached pair per generator and clean image. VisA memory uses the same three intervention families under the uniform q50 policy. Reconstruction samples one of the three interventions uniformly for each training example. Donors are same-category, and Real-IAD additionally enforces same camera view.

\section{Pixel-Preserved Patch Construction}
\label{app:mask}
The change mask in Eq.~\eqref{eq:change} is evaluated in signed arithmetic to avoid uint8 wraparound. Let
\[
A_P=P(A),\qquad D_P=P(D),\qquad C=A_P\lor D_P,
\]
where $P$ is adaptive max pooling to the $28\times28$ token grid. At the final $392\times392$ crop, each token corresponds to a $14\times14$ pixel footprint. The strict retained set is
\[
R_{\mathrm{strict}}=\operatorname{MaxPool}_{2r+1}(A_P)\land\neg C,\qquad r=2.
\]
The dilation is applied to $A_P$. A token changed outside the nominal insertion is excluded through $D_P$ but does not expand the ring.

For the nominal-mask-only control, the retained set is
\[
R_{\mathrm{nominal}}=\operatorname{MaxPool}_{2r+1}(A_P)\land\neg A_P.
\]
This set can retain tokens intersecting detected RGB changes outside the nominal mask. Main memory candidates and reconstruction targets are restricted to $R_{\mathrm{strict}}$. The synthetic insertion remains part of the model input; the main reconstruction configuration does not mask its attention keys/values.

\section{Memory Branch Details}
\label{app:memory}
The memory branch uses frozen plain DINOv2 ViT-B/14 \citep{oquab2024dinov2} without register tokens and extracts layers $[-3,-6,-8,-9]$. Each 768-dimensional token is $\ell_2$-normalized and mapped through a fixed Gaussian projection (seed 42) to 512 dimensions. The projected feature is not normalized again.

For each layer, five banks are selected independently with approximate greedy $k$-center. Bank $b$ uses seed $s+b$, a separate random 192-dimensional coreset projection, ten random starts, and target size
\[
B=\max\{1,\lfloor\rho N\rfloor\}.
\]
Caches and final banks are stored in FP16, while candidate distances, coreset selection, and inference distances are evaluated in FP32; TF32 is enabled.

\subsection{Candidate score}
Candidate discovery uses the bank-wise nearest Euclidean distance
\[
d_\ell(z)=\operatorname{median}_{b=1}^{5}\min_{m\in M_{\ell,b}}\|z-m\|_2,
\qquad
s_{\mathrm{cand}}(z)=\frac14\sum_\ell d_\ell(z).
\]
For a clean/synthetic pair at the same position,
\[
\Delta_p=s_{\mathrm{cand}}(\tilde z_p)-s_{\mathrm{cand}}(z_p).
\]
Quantiles are computed separately within each intervention type. The uniform policy selects a shared four-layer position inside $R_{\mathrm{strict}}$ when
\[
s_{\mathrm{cand}}(z_p)\le Q_{0.50},\qquad \Delta_p\ge Q_{0.90},\qquad \Delta_p>0.
\]
Selected altered-context features are concatenated with all clean candidates, and the same $k$-center selection procedure selects exactly $B$ final references per bank. MVTec/VisA use $\rho=0.05$ and Real-IAD uses $\rho=0.01$.

\subsection{Test-time readout}
The candidate-selection score is distinct from the deployed ProCon readout \citep{chae2026procon}. Inference uses five nearest anchors per bank, soft reconstruction with a temperature derived from the median squared neighbor distance on at most 512 query tokens, median fusion across five banks, and averaging across four layers. The clean baseline and BoundarySupport share the same readout.

\section{Reconstruction Branch Details}
\label{app:reconstruction}
The reconstruction branch uses frozen \texttt{dinov2reg\_vit\_base\_14}. Encoder blocks 2--9 provide eight targets, averaged into two groups over blocks 2--5 and 6--9. Trainable components are a 768--3072--768 bottleneck with dropout 0.2 and eight 12-head linear-attention decoder blocks.

Each category is trained for 5,000 iterations with batch size 16 using StableAdamW: learning rate $2\times10^{-3}$, betas $(0.9,0.999)$, weight decay $10^{-4}$, AMSGrad, gradient clipping at 0.1, 100 warm-up iterations, and cosine decay to $2\times10^{-4}$. Training is FP32. The clean hard-mining schedule is
\[
p_t=\min(0.9t/1000,0.9),
\]
with gradients below the token-residual threshold scaled by 0.1.

For fused encoder target $E_j(v)_p$ and decoder output $G_j(v)_p$, the pixel-preserved-patch objective is
\[
\mathcal L_R=\frac12\sum_{j=1}^{2}\operatorname{mean}_{p\in R_{\mathrm{strict}}}
\left[1-\cos\left(\operatorname{sg}(E_j(\tilde x)_p),G_j(\tilde x)_p\right)\right],
\]
and
\[
\mathcal L=\mathcal L_{\mathrm{base}}+0.05\mathcal L_R.
\]
The two terms are backpropagated separately before one clipping and optimizer step, which is gradient-equivalent to their sum. The main configuration uses $r=2$, auxiliary weight 0.05, and no core-key mask. Testing uses the baseline scoring path without synthesis or masks.

\clearpage
\section{Full Dataset-Level Metrics}
\label{app:fullmetrics}
\begin{table}[!htbp]
\centering
\scriptsize
\setlength{\tabcolsep}{3pt}
\caption{Seed-0 category-macro metrics. Three-seed paired changes are reported in Table~\ref{tab:main} and Appendix~\ref{app:seeds}.}
\resizebox{\textwidth}{!}{%
\begin{tabular}{lllrrrrrrr}
\toprule
Branch & Dataset & Variant & I-AUROC & I-AP & I-F1 & P-AUROC & P-AP & P-F1 & AUPRO \\
\midrule
Memory & MVTec & Baseline & 99.76 & 99.91 & 99.38 & 98.80 & 73.34 & 71.04 & 96.18 \\
Memory & MVTec & BoundarySupport & 99.73 & 99.90 & 99.31 & 98.86 & 75.60 & 72.33 & 96.39 \\
Memory & VisA & Baseline & 99.17 & 99.28 & 97.50 & 99.08 & 52.23 & 54.71 & 97.06 \\
Memory & VisA & BoundarySupport & 99.19 & 99.30 & 97.54 & 99.07 & 54.05 & 55.75 & 97.03 \\
Memory & Real-IAD & Baseline & 94.89 & 94.33 & 89.00 & 99.17 & 52.02 & 53.11 & 97.60 \\
Memory & Real-IAD & BoundarySupport & 94.79 & 94.11 & 88.76 & 99.16 & 53.78 & 54.42 & 97.54 \\
Recon. & MVTec & Baseline & 99.75 & 99.91 & 99.24 & 98.35 & 68.95 & 68.79 & 94.69 \\
Recon. & MVTec & BoundarySupport & 99.68 & 99.86 & 99.07 & 98.53 & 73.89 & 71.89 & 95.20 \\
Recon. & VisA & Baseline & 98.93 & 99.04 & 96.80 & 98.97 & 50.70 & 54.29 & 94.77 \\
Recon. & VisA & BoundarySupport & 98.99 & 99.12 & 96.69 & 98.80 & 52.79 & 55.21 & 94.82 \\
Recon. & Real-IAD & Baseline & 94.27 & 93.55 & 88.15 & 99.29 & 49.46 & 52.06 & 96.24 \\
Recon. & Real-IAD & BoundarySupport & 93.75 & 92.87 & 87.26 & 99.26 & 53.66 & 54.62 & 96.40 \\
\bottomrule
\end{tabular}}
\end{table}

\FloatBarrier
\section{Repeated Runs and Sign Consistency}
\label{app:seeds}
\begin{table}[h]
\centering
\small
\caption{P-AP over three paired seeds (0, 1, 2).}
\begin{tabular}{llrrrr}
\toprule
Branch & Dataset & Baseline & BoundarySupport & Paired $\Delta$ & Positive pairs \\
\midrule
Memory & MVTec & $73.35\pm0.01$ & $75.56\pm0.03$ & $+2.21\pm0.04$ & 45/45 \\
Memory & VisA & $52.23\pm0.02$ & $54.05\pm0.03$ & $+1.82\pm0.03$ & 36/36 \\
Memory & Real-IAD & $52.02\pm0.02$ & $53.78\pm0.03$ & $+1.76\pm0.04$ & 90/90 \\
Recon. & MVTec & $68.96\pm0.11$ & $73.98\pm0.11$ & $+5.01\pm0.07$ & 45/45 \\
Recon. & VisA & $50.93\pm0.26$ & $53.20\pm0.39$ & $+2.27\pm0.38$ & 29/36 \\
Recon. & Real-IAD & $49.46\pm0.15$ & $53.66\pm0.18$ & $+4.20\pm0.12$ & 84/90 \\
\bottomrule
\end{tabular}
\end{table}
Memory is positive in all 171 category--seed pairs, while reconstruction is positive in 158/171; overall, P-AP increases in 329/342 pairs. These counts describe sign consistency. The reported mean$\pm$sample standard deviation is computed from the three seed-level category macros.

\FloatBarrier
\section{Mechanism Controls}
\label{app:controls}
All controls in this section use seed 0. Controls 3 and 4 cover all 15 MVTec and 12 VisA categories. Control 5 covers all 15 MVTec categories.

\subsection{Control 3: Reconstruction target identity}
At positions $p\in R_{\mathrm{strict}}$, BoundarySupport uses
\[
T_p^{\mathrm{altered}}=\operatorname{sg}[E(\tilde x)_p],
\]
whereas the clean-target control keeps the same synthetic input and ring but uses
\[
T_p^{\mathrm{clean}}=\operatorname{sg}[E(x)_p].
\]
The synthetic input, $R_{\mathrm{strict}}$, frozen encoder, bottleneck/decoder, auxiliary weight, training iterations, and seed are otherwise identical.

\begin{table}[h]
\centering
\small
\caption{Control 3 P-AP: target identity with synthetic input and preserved positions fixed.}
\begin{tabular}{lrrrr}
\toprule
Dataset & Clean-view target & Altered-context target & Altered $-$ clean & Category wins \\
\midrule
MVTec & 61.82 & 73.89 & +12.07 & 15/15 \\
VisA & 44.34 & 52.79 & +8.45 & 11/12 \\
\bottomrule
\end{tabular}
\end{table}
On MVTec, the clean target is 7.13 points below baseline and decreases P-AP in 15/15 categories; the altered-context target is 4.94 points above baseline and improves 15/15. The category-median altered-minus-clean difference is 11.84 points. On VisA, the clean target is 6.36 points below baseline; the altered-context target is 2.09 points above baseline, with an altered-minus-clean difference of 8.45 points and 11/12 category wins. AUPRO changes relative to baseline are $-0.55/+0.51$ (clean/altered) on MVTec and $-0.17/+0.05$ on VisA. I-AUROC changes remain within 0.08 point in these target variants.

\subsection{Control 4: Memory feature identity}
Let $\mathcal P_{\mathrm{sel}}$ be the positions selected by the same candidate rule. We keep these positions, the clean candidate pool, absolute bank budget, joint $k$-center selection, bank count, projections, inference, and seed fixed, and compare
\[
\mathcal Z_{\mathrm{clean-pos}}=\{F(x)_p:p\in\mathcal P_{\mathrm{sel}}\},\qquad
\mathcal Z_{\mathrm{altered}}=\{F(\tilde x)_p:p\in\mathcal P_{\mathrm{sel}}\}.
\]
Only feature values differ before the same joint $k$-center selection constructs the final memory.

\begin{table}[h]
\centering
\small
\caption{Control 4 P-AP: stored-feature identity with selected positions and bank construction fixed. Deltas are computed from unrounded values.}
\begin{tabular}{lrrrr}
\toprule
Dataset & Clean-view feature & Altered-context feature & $\Delta$P-AP & Category wins \\
\midrule
MVTec & 73.36 & 75.60 & +2.23 & 15/15 \\
VisA & 52.28 & 54.11 & +1.84 & 11/12 \\
\bottomrule
\end{tabular}
\end{table}
The category-median P-AP differences are +1.78 on MVTec and +1.14 on VisA. MVTec AUPRO changes from 96.21 to 96.39, with 13/15 category wins, while I-AUROC is 99.74 versus 99.73. On VisA, AUPRO is 97.08 versus 97.03 and I-AUROC 99.17 versus 99.16. Across the two datasets, altered-context features yield higher P-AP in 26 of 27 categories. The only VisA reversal is candle at -0.23 point.

\subsection{Control 5: Distance-resolved final memory scores}
For each anomalous MVTec test image, normal patches are partitioned by patch-grid distance to the nearest anomalous patch: near $0<d\le2$, mid $2<d\le4$, and far $d>4$. Defect patches and normal patches from clean images are aggregated separately. For each patch,
\[
\Delta s(p)=s_{\mathrm{BoundarySupport}}(p)-s_{\mathrm{baseline}}(p).
\]

\begin{table}[h]
\centering
\small
\caption{Control 5 mean MVTec memory score change. Values are $100\Delta s$.}
\begin{tabular}{lr}
\toprule
Spatial cohort & Mean $100\Delta s$ \\
\midrule
Defect patches & -0.61 \\
Near-defect normal & -0.37 \\
Mid-distance normal & +0.11 \\
Far normal & +0.19 \\
Clean-image normal & +0.21 \\
\bottomrule
\end{tabular}
\end{table}
Near-minus-far and near-minus-mid contrasts are negative in 15/15 categories. Their category-median contrasts on the $100\Delta s$ scale are -0.51 and -0.43, respectively. Thus the final memory changes normal responses more strongly next to real defects than at mid or far distances.

\subsection{Additional reconstruction-object controls}
The following controls change the supervised patch set or objective while retaining the same reconstruction architecture and auxiliary weight. The corresponding clean-view patch applies the auxiliary objective to matched clean-view positions. The nominal-ring control omits detected-RGB-change exclusion. The synthetic-core control applies a margin objective to the synthetic core.

\begin{table}[!htbp]
\centering
\scriptsize
\setlength{\tabcolsep}{2.5pt}
\caption{Seed-0 full-dataset reconstruction-object controls.}
\resizebox{\textwidth}{!}{%
\begin{tabular}{llrrrrrrrr}
\toprule
Dataset & Supervised object & $\Delta$I-AUROC & $\Delta$I-AP & $\Delta$I-F1 & $\Delta$P-AUROC & $\Delta$P-AP & $\Delta$P-F1 & $\Delta$AUPRO & P-AP wins \\
\midrule
MVTec & Pixel-preserved surrounding patch & -0.07 & -0.05 & -0.18 & +0.18 & +4.94 & +3.10 & +0.51 & 15/15 \\
MVTec & Corresponding clean-view patch & -0.02 & 0.00 & +0.04 & -0.01 & -0.26 & -0.15 & -0.01 & 7/15 \\
MVTec & Patch outside nominal mask & -0.23 & -0.20 & -0.71 & -0.03 & +0.70 & +0.99 & -0.09 & 8/15 \\
MVTec & Synthetic core & -2.11 & -1.44 & -0.75 & -20.86 & -52.28 & -45.18 & -49.13 & 0/15 \\
VisA & Pixel-preserved surrounding patch & +0.06 & +0.08 & -0.11 & -0.17 & +2.09 & +0.92 & +0.05 & 9/12 \\
VisA & Corresponding clean-view patch & -0.27 & -0.32 & -0.65 & -0.07 & -0.30 & -0.16 & -0.28 & 4/12 \\
VisA & Patch outside nominal mask & -0.14 & -0.13 & -0.30 & -0.33 & -2.40 & -1.31 & +0.09 & 5/12 \\
VisA & Synthetic core & -1.10 & -1.67 & -0.90 & -9.44 & -42.67 & -40.07 & -42.02 & 0/12 \\
\bottomrule
\end{tabular}}
\end{table}

\FloatBarrier
\section{Generator and Hyperparameter Analysis}
\label{app:ablations}
These seed-0 reconstruction ablations use ten development categories: the five MVTec textures and VisA candle, capsules, chewinggum, macaroni2, and pipe-fryum.

\begin{table}[!htbp]
\centering
\scriptsize
\setlength{\tabcolsep}{3pt}
\caption{Paired changes on the ten-category development subset.}
\resizebox{\textwidth}{!}{%
\begin{tabular}{lrrrrrrr}
\toprule
Setting & $\Delta$I-AUROC & $\Delta$I-AP & $\Delta$I-F1 & $\Delta$P-AUROC & $\Delta$P-AP & $\Delta$P-F1 & $\Delta$AUPRO \\
\midrule
Three interventions, $r=2$, $\lambda=0.05$ & +0.01 & +0.01 & -0.21 & +0.23 & +6.99 & +4.18 & +0.80 \\
Scar only & +0.01 & +0.01 & -0.04 & +0.26 & +8.29 & +4.58 & +0.80 \\
Poisson paste only & +0.12 & +0.07 & +0.40 & +0.18 & +4.45 & +2.93 & +1.09 \\
Local warp + paste only & +0.04 & +0.02 & +0.05 & +0.15 & +4.60 & +2.85 & +0.70 \\
$r=1$ & +0.05 & +0.03 & -0.04 & +0.25 & +7.55 & +4.31 & +0.99 \\
$r=3$ & 0.00 & -0.01 & -0.04 & +0.23 & +6.91 & +4.11 & +1.21 \\
$\lambda=0.025$ & +0.04 & +0.03 & +0.02 & +0.23 & +6.66 & +3.80 & +1.17 \\
$\lambda=0.1$ & +0.05 & +0.03 & +0.02 & +0.23 & +7.31 & +4.12 & +1.01 \\
Core-key masking & +0.03 & +0.01 & -0.17 & +0.23 & +6.87 & +4.23 & +0.92 \\
\bottomrule
\end{tabular}}
\end{table}
All three intervention families are positive on this subset. Positive P-AP also persists for $r\in\{1,2,3\}$ and $\lambda\in\{0.025,0.05,0.1\}$.

\FloatBarrier
\section{Category-Level Results}
\label{app:categories}
All category tables below use seed 0 and the 0--100 scale. Cells with arrows report baseline $\rightarrow$ BoundarySupport unless noted otherwise.
\begin{table}[!htbp]
\centering
\scriptsize
\setlength{\tabcolsep}{3pt}
\caption{Category-level seed-0 results for MVTec Memory. Cells with arrows report baseline $\rightarrow$ BoundarySupport.}
\resizebox{\textwidth}{!}{%
\begin{tabular}{lrrrrrrr}
\toprule
Category & I-AUROC & I-AP & I-F1 & P-AUROC & P-AP & P-F1 & AUPRO \\
\midrule
bottle & 100.00$\rightarrow$100.00 & 100.00$\rightarrow$100.00 & 100.00$\rightarrow$100.00 & 99.20$\rightarrow$99.26 & 89.02$\rightarrow$90.79 & 82.61$\rightarrow$83.12 & 97.33$\rightarrow$97.54 \\
cable & 99.81$\rightarrow$99.78 & 99.86$\rightarrow$99.85 & 98.14$\rightarrow$98.14 & 98.66$\rightarrow$98.66 & 76.70$\rightarrow$77.68 & 72.22$\rightarrow$72.54 & 95.66$\rightarrow$95.71 \\
capsule & 98.75$\rightarrow$98.75 & 99.67$\rightarrow$99.67 & 98.48$\rightarrow$98.48 & 98.98$\rightarrow$98.99 & 64.40$\rightarrow$65.03 & 61.17$\rightarrow$61.53 & 96.66$\rightarrow$96.70 \\
carpet & 100.00$\rightarrow$100.00 & 100.00$\rightarrow$100.00 & 100.00$\rightarrow$100.00 & 99.58$\rightarrow$99.61 & 80.07$\rightarrow$81.90 & 76.78$\rightarrow$77.68 & 98.61$\rightarrow$98.70 \\
grid & 100.00$\rightarrow$100.00 & 100.00$\rightarrow$100.00 & 100.00$\rightarrow$100.00 & 99.63$\rightarrow$99.67 & 63.74$\rightarrow$67.48 & 61.84$\rightarrow$64.04 & 98.78$\rightarrow$98.91 \\
hazelnut & 100.00$\rightarrow$100.00 & 100.00$\rightarrow$100.00 & 100.00$\rightarrow$100.00 & 99.57$\rightarrow$99.59 & 81.64$\rightarrow$82.75 & 77.43$\rightarrow$78.43 & 98.56$\rightarrow$98.65 \\
leather & 100.00$\rightarrow$100.00 & 100.00$\rightarrow$100.00 & 100.00$\rightarrow$100.00 & 99.59$\rightarrow$99.68 & 61.97$\rightarrow$68.55 & 60.39$\rightarrow$65.28 & 98.63$\rightarrow$98.95 \\
metal\_nut & 100.00$\rightarrow$100.00 & 100.00$\rightarrow$100.00 & 100.00$\rightarrow$100.00 & 97.90$\rightarrow$97.94 & 83.06$\rightarrow$83.62 & 87.67$\rightarrow$87.75 & 93.05$\rightarrow$93.19 \\
pill & 99.55$\rightarrow$99.39 & 99.91$\rightarrow$99.89 & 99.21$\rightarrow$99.21 & 97.42$\rightarrow$97.51 & 75.61$\rightarrow$76.89 & 70.25$\rightarrow$70.79 & 92.81$\rightarrow$93.02 \\
screw & 98.66$\rightarrow$98.52 & 99.46$\rightarrow$99.41 & 97.59$\rightarrow$96.62 & 99.46$\rightarrow$99.46 & 65.89$\rightarrow$66.13 & 61.42$\rightarrow$61.49 & 98.37$\rightarrow$98.35 \\
tile & 100.00$\rightarrow$100.00 & 100.00$\rightarrow$100.00 & 100.00$\rightarrow$100.00 & 98.51$\rightarrow$98.82 & 76.69$\rightarrow$81.29 & 79.68$\rightarrow$82.02 & 95.05$\rightarrow$96.06 \\
toothbrush & 100.00$\rightarrow$100.00 & 100.00$\rightarrow$100.00 & 100.00$\rightarrow$100.00 & 99.32$\rightarrow$99.39 & 63.00$\rightarrow$66.02 & 69.29$\rightarrow$70.86 & 97.73$\rightarrow$97.96 \\
transistor & 99.94$\rightarrow$99.89 & 99.89$\rightarrow$99.78 & 98.31$\rightarrow$98.25 & 96.55$\rightarrow$96.56 & 66.84$\rightarrow$67.48 & 64.09$\rightarrow$64.40 & 89.40$\rightarrow$89.50 \\
wood & 99.66$\rightarrow$99.66 & 99.86$\rightarrow$99.86 & 98.95$\rightarrow$98.95 & 99.03$\rightarrow$99.16 & 79.57$\rightarrow$83.14 & 72.66$\rightarrow$75.34 & 96.78$\rightarrow$97.21 \\
zipper & 100.00$\rightarrow$100.00 & 100.00$\rightarrow$100.00 & 100.00$\rightarrow$100.00 & 98.57$\rightarrow$98.59 & 71.83$\rightarrow$75.20 & 68.04$\rightarrow$69.71 & 95.37$\rightarrow$95.44 \\
\bottomrule
\end{tabular}}
\end{table}

\begin{table}[!htbp]
\centering
\scriptsize
\setlength{\tabcolsep}{3pt}
\caption{Category-level seed-0 results for VisA Memory, uniform q50. Cells with arrows report baseline $\rightarrow$ BoundarySupport.}
\begin{tabular}{lrr}
\toprule
Category & Baseline P-AP & Uniform $\Delta$P-AP \\
\midrule
candle & 46.04 & +0.33 \\
capsules & 68.93 & +1.14 \\
cashew & 69.10 & +2.75 \\
chewinggum & 73.90 & +6.19 \\
fryum & 46.32 & +0.42 \\
macaroni1 & 27.64 & +2.47 \\
macaroni2 & 20.92 & +0.93 \\
pcb1 & 87.45 & +0.98 \\
pcb2 & 34.38 & +1.66 \\
pcb3 & 39.35 & +0.58 \\
pcb4 & 52.28 & +0.24 \\
pipe\_fryum & 60.46 & +4.16 \\
\textbf{Macro} & 52.23 & +1.82 \\
\bottomrule
\end{tabular}
\end{table}

\begin{table}[!htbp]
\centering
\scriptsize
\setlength{\tabcolsep}{3pt}
\caption{Category-level seed-0 results for Real-IAD Memory. Cells with arrows report baseline $\rightarrow$ BoundarySupport.}
\resizebox{\textwidth}{!}{%
\begin{tabular}{lrrrrrrr}
\toprule
Category & I-AUROC & I-AP & I-F1 & P-AUROC & P-AP & P-F1 & AUPRO \\
\midrule
audiojack & 96.51$\rightarrow$96.62 & 94.91$\rightarrow$95.00 & 87.75$\rightarrow$87.44 & 99.83$\rightarrow$99.84 & 60.97$\rightarrow$68.74 & 61.49$\rightarrow$64.60 & 99.44$\rightarrow$99.46 \\
bottle\_cap & 96.27$\rightarrow$96.29 & 96.54$\rightarrow$96.61 & 90.62$\rightarrow$90.91 & 99.64$\rightarrow$99.60 & 35.04$\rightarrow$36.86 & 38.59$\rightarrow$40.24 & 98.80$\rightarrow$98.68 \\
button\_battery & 91.70$\rightarrow$91.29 & 93.99$\rightarrow$93.77 & 87.81$\rightarrow$87.84 & 99.62$\rightarrow$99.62 & 70.06$\rightarrow$70.51 & 66.73$\rightarrow$67.07 & 98.73$\rightarrow$98.75 \\
end\_cap & 91.12$\rightarrow$90.72 & 90.85$\rightarrow$90.58 & 88.39$\rightarrow$88.01 & 99.38$\rightarrow$99.34 & 14.10$\rightarrow$14.77 & 26.00$\rightarrow$26.64 & 98.01$\rightarrow$97.86 \\
eraser & 95.25$\rightarrow$95.14 & 94.39$\rightarrow$94.32 & 86.46$\rightarrow$86.05 & 99.79$\rightarrow$99.79 & 49.00$\rightarrow$50.48 & 51.47$\rightarrow$52.09 & 99.30$\rightarrow$99.31 \\
fire\_hood & 96.09$\rightarrow$95.72 & 92.27$\rightarrow$91.51 & 87.28$\rightarrow$86.55 & 99.66$\rightarrow$99.63 & 51.23$\rightarrow$52.77 & 51.29$\rightarrow$51.81 & 98.88$\rightarrow$98.81 \\
mint & 77.24$\rightarrow$77.85 & 85.89$\rightarrow$86.27 & 76.47$\rightarrow$76.99 & 99.27$\rightarrow$99.20 & 44.12$\rightarrow$45.42 & 47.28$\rightarrow$48.22 & 97.86$\rightarrow$97.72 \\
mounts & 97.55$\rightarrow$97.53 & 96.45$\rightarrow$96.45 & 90.53$\rightarrow$90.15 & 99.74$\rightarrow$99.75 & 50.14$\rightarrow$51.82 & 51.36$\rightarrow$52.36 & 99.15$\rightarrow$99.18 \\
pcb & 96.92$\rightarrow$96.97 & 98.18$\rightarrow$98.19 & 93.02$\rightarrow$93.18 & 99.82$\rightarrow$99.82 & 61.81$\rightarrow$63.05 & 60.21$\rightarrow$60.93 & 99.40$\rightarrow$99.40 \\
phone\_battery & 97.24$\rightarrow$97.18 & 97.46$\rightarrow$97.39 & 91.57$\rightarrow$91.85 & 99.89$\rightarrow$99.90 & 67.31$\rightarrow$69.76 & 63.20$\rightarrow$65.56 & 99.65$\rightarrow$99.65 \\
plastic\_nut & 95.92$\rightarrow$95.89 & 88.24$\rightarrow$87.92 & 84.58$\rightarrow$82.91 & 99.87$\rightarrow$99.84 & 50.41$\rightarrow$52.57 & 48.66$\rightarrow$50.31 & 99.56$\rightarrow$99.49 \\
plastic\_plug & 96.47$\rightarrow$96.30 & 95.85$\rightarrow$95.61 & 88.71$\rightarrow$88.39 & 99.54$\rightarrow$99.54 & 37.21$\rightarrow$39.43 & 40.56$\rightarrow$43.29 & 98.74$\rightarrow$98.77 \\
porcelain\_doll & 95.39$\rightarrow$95.19 & 92.66$\rightarrow$92.48 & 85.50$\rightarrow$85.23 & 99.53$\rightarrow$99.52 & 32.10$\rightarrow$34.04 & 37.19$\rightarrow$38.56 & 98.77$\rightarrow$98.76 \\
regulator & 96.49$\rightarrow$96.41 & 90.57$\rightarrow$90.60 & 88.19$\rightarrow$87.30 & 99.97$\rightarrow$99.97 & 73.60$\rightarrow$74.17 & 69.13$\rightarrow$69.82 & 99.91$\rightarrow$99.91 \\
rolled\_strip\_base & 99.69$\rightarrow$99.61 & 99.84$\rightarrow$99.80 & 99.02$\rightarrow$98.93 & 99.78$\rightarrow$99.80 & 44.67$\rightarrow$47.19 & 45.64$\rightarrow$47.64 & 99.27$\rightarrow$99.32 \\
sim\_card\_set & 99.03$\rightarrow$99.01 & 99.32$\rightarrow$99.30 & 96.23$\rightarrow$96.22 & 99.88$\rightarrow$99.88 & 72.55$\rightarrow$74.64 & 65.77$\rightarrow$67.77 & 99.59$\rightarrow$99.61 \\
switch & 99.14$\rightarrow$99.09 & 99.55$\rightarrow$99.53 & 97.59$\rightarrow$97.59 & 99.38$\rightarrow$99.38 & 65.09$\rightarrow$65.18 & 63.88$\rightarrow$64.05 & 97.94$\rightarrow$97.94 \\
tape & 99.67$\rightarrow$99.66 & 99.48$\rightarrow$99.47 & 96.63$\rightarrow$96.85 & 99.90$\rightarrow$99.91 & 58.45$\rightarrow$62.31 & 56.26$\rightarrow$59.14 & 99.68$\rightarrow$99.70 \\
terminalblock & 99.25$\rightarrow$99.29 & 99.54$\rightarrow$99.57 & 98.07$\rightarrow$98.21 & 99.91$\rightarrow$99.91 & 62.46$\rightarrow$65.15 & 61.38$\rightarrow$63.59 & 99.69$\rightarrow$99.71 \\
toothbrush & 92.80$\rightarrow$92.77 & 95.99$\rightarrow$95.98 & 88.89$\rightarrow$88.44 & 99.23$\rightarrow$99.22 & 44.61$\rightarrow$44.86 & 51.49$\rightarrow$51.75 & 97.42$\rightarrow$97.41 \\
toy & 90.76$\rightarrow$91.09 & 95.76$\rightarrow$95.92 & 86.85$\rightarrow$87.33 & 88.59$\rightarrow$88.53 & 24.99$\rightarrow$25.58 & 35.41$\rightarrow$35.62 & 69.69$\rightarrow$68.73 \\
toy\_brick & 84.61$\rightarrow$83.67 & 85.22$\rightarrow$83.60 & 75.98$\rightarrow$74.85 & 97.79$\rightarrow$97.74 & 52.78$\rightarrow$54.58 & 53.93$\rightarrow$55.57 & 93.54$\rightarrow$93.37 \\
transistor1 & 97.39$\rightarrow$97.48 & 98.47$\rightarrow$98.53 & 94.87$\rightarrow$95.04 & 99.52$\rightarrow$99.49 & 46.87$\rightarrow$47.98 & 46.94$\rightarrow$47.60 & 98.39$\rightarrow$98.32 \\
u\_block & 96.19$\rightarrow$95.99 & 93.37$\rightarrow$92.87 & 89.45$\rightarrow$89.08 & 99.81$\rightarrow$99.80 & 62.18$\rightarrow$65.50 & 60.11$\rightarrow$62.24 & 99.39$\rightarrow$99.37 \\
usb & 97.46$\rightarrow$97.36 & 96.90$\rightarrow$96.83 & 92.54$\rightarrow$92.68 & 99.83$\rightarrow$99.84 & 52.29$\rightarrow$52.45 & 53.75$\rightarrow$54.33 & 99.45$\rightarrow$99.46 \\
usb\_adaptor & 93.42$\rightarrow$93.08 & 91.24$\rightarrow$90.88 & 84.97$\rightarrow$85.09 & 99.81$\rightarrow$99.81 & 31.58$\rightarrow$33.31 & 35.70$\rightarrow$38.34 & 99.37$\rightarrow$99.38 \\
vcpill & 95.63$\rightarrow$95.91 & 94.10$\rightarrow$94.38 & 85.93$\rightarrow$86.01 & 98.53$\rightarrow$98.56 & 72.44$\rightarrow$72.92 & 70.44$\rightarrow$70.61 & 95.97$\rightarrow$96.01 \\
wooden\_beads & 89.56$\rightarrow$89.58 & 92.64$\rightarrow$92.41 & 83.59$\rightarrow$83.49 & 99.00$\rightarrow$98.98 & 49.79$\rightarrow$51.04 & 52.40$\rightarrow$53.44 & 97.26$\rightarrow$97.17 \\
woodstick & 92.24$\rightarrow$91.19 & 80.34$\rightarrow$77.68 & 73.59$\rightarrow$71.43 & 99.56$\rightarrow$99.52 & 60.42$\rightarrow$61.74 & 60.61$\rightarrow$62.27 & 98.52$\rightarrow$98.40 \\
zipper & 99.83$\rightarrow$99.81 & 99.92$\rightarrow$99.90 & 99.00$\rightarrow$98.91 & 98.89$\rightarrow$98.91 & 62.41$\rightarrow$64.70 & 66.32$\rightarrow$67.16 & 96.52$\rightarrow$96.61 \\
\bottomrule
\end{tabular}}
\end{table}

\begin{table}[!htbp]
\centering
\scriptsize
\setlength{\tabcolsep}{3pt}
\caption{Category-level seed-0 results for MVTec Reconstruction. Cells with arrows report baseline $\rightarrow$ BoundarySupport.}
\resizebox{\textwidth}{!}{%
\begin{tabular}{lrrrrrrr}
\toprule
Category & I-AUROC & I-AP & I-F1 & P-AUROC & P-AP & P-F1 & AUPRO \\
\midrule
bottle & 100.00$\rightarrow$100.00 & 100.00$\rightarrow$100.00 & 100.00$\rightarrow$100.00 & 99.14$\rightarrow$99.19 & 88.88$\rightarrow$91.02 & 83.43$\rightarrow$84.20 & 97.16$\rightarrow$97.59 \\
cable & 99.94$\rightarrow$100.00 & 99.97$\rightarrow$100.00 & 99.46$\rightarrow$100.00 & 98.25$\rightarrow$98.28 & 76.09$\rightarrow$77.57 & 73.70$\rightarrow$74.75 & 93.67$\rightarrow$93.30 \\
capsule & 98.96$\rightarrow$98.44 & 99.77$\rightarrow$99.65 & 99.08$\rightarrow$99.08 & 98.68$\rightarrow$98.59 & 60.32$\rightarrow$61.60 & 58.54$\rightarrow$59.16 & 97.74$\rightarrow$97.30 \\
carpet & 99.96$\rightarrow$99.84 & 99.99$\rightarrow$99.95 & 99.44$\rightarrow$99.44 & 99.34$\rightarrow$99.42 & 68.56$\rightarrow$74.44 & 71.60$\rightarrow$75.10 & 97.36$\rightarrow$97.96 \\
grid & 99.92$\rightarrow$99.92 & 99.97$\rightarrow$99.97 & 99.13$\rightarrow$99.13 & 99.44$\rightarrow$99.55 & 57.34$\rightarrow$65.98 & 59.84$\rightarrow$64.83 & 97.52$\rightarrow$97.89 \\
hazelnut & 100.00$\rightarrow$100.00 & 100.00$\rightarrow$100.00 & 100.00$\rightarrow$100.00 & 99.46$\rightarrow$99.57 & 82.44$\rightarrow$85.67 & 77.52$\rightarrow$80.50 & 96.63$\rightarrow$97.50 \\
leather & 100.00$\rightarrow$100.00 & 100.00$\rightarrow$100.00 & 100.00$\rightarrow$100.00 & 99.34$\rightarrow$99.63 & 52.80$\rightarrow$65.53 & 54.14$\rightarrow$66.06 & 98.24$\rightarrow$98.60 \\
metal\_nut & 100.00$\rightarrow$100.00 & 100.00$\rightarrow$100.00 & 100.00$\rightarrow$100.00 & 96.99$\rightarrow$96.99 & 80.23$\rightarrow$82.05 & 86.16$\rightarrow$86.52 & 94.78$\rightarrow$95.09 \\
pill & 99.32$\rightarrow$99.48 & 99.88$\rightarrow$99.91 & 98.93$\rightarrow$98.93 & 97.43$\rightarrow$97.70 & 70.94$\rightarrow$75.15 & 68.29$\rightarrow$70.50 & 96.30$\rightarrow$97.54 \\
screw & 98.93$\rightarrow$98.61 & 99.64$\rightarrow$99.53 & 97.07$\rightarrow$96.23 & 99.67$\rightarrow$99.64 & 62.42$\rightarrow$62.51 & 60.22$\rightarrow$59.88 & 98.44$\rightarrow$98.30 \\
tile & 100.00$\rightarrow$100.00 & 100.00$\rightarrow$100.00 & 100.00$\rightarrow$100.00 & 97.17$\rightarrow$98.51 & 69.66$\rightarrow$80.20 & 72.66$\rightarrow$79.96 & 84.78$\rightarrow$89.68 \\
toothbrush & 100.00$\rightarrow$100.00 & 100.00$\rightarrow$100.00 & 100.00$\rightarrow$100.00 & 99.00$\rightarrow$99.23 & 55.12$\rightarrow$62.83 & 64.65$\rightarrow$68.94 & 95.56$\rightarrow$96.23 \\
transistor & 99.75$\rightarrow$99.33 & 99.63$\rightarrow$98.99 & 97.56$\rightarrow$95.24 & 94.85$\rightarrow$94.78 & 64.97$\rightarrow$65.01 & 61.74$\rightarrow$60.74 & 81.84$\rightarrow$80.11 \\
wood & 99.56$\rightarrow$99.65 & 99.86$\rightarrow$99.89 & 98.36$\rightarrow$98.36 & 97.49$\rightarrow$97.99 & 70.61$\rightarrow$79.80 & 68.17$\rightarrow$72.67 & 93.53$\rightarrow$94.18 \\
zipper & 99.95$\rightarrow$99.97 & 99.99$\rightarrow$99.99 & 99.58$\rightarrow$99.58 & 99.01$\rightarrow$98.92 & 73.83$\rightarrow$79.03 & 71.15$\rightarrow$74.57 & 96.77$\rightarrow$96.67 \\
\bottomrule
\end{tabular}}
\end{table}

\begin{table}[!htbp]
\centering
\scriptsize
\setlength{\tabcolsep}{3pt}
\caption{Category-level seed-0 results for VisA Reconstruction. Cells with arrows report baseline $\rightarrow$ BoundarySupport.}
\resizebox{\textwidth}{!}{%
\begin{tabular}{lrrrrrrr}
\toprule
Category & I-AUROC & I-AP & I-F1 & P-AUROC & P-AP & P-F1 & AUPRO \\
\midrule
candle & 98.14$\rightarrow$98.11 & 98.20$\rightarrow$98.13 & 93.53$\rightarrow$93.60 & 99.43$\rightarrow$99.41 & 43.30$\rightarrow$43.60 & 48.85$\rightarrow$50.79 & 95.50$\rightarrow$95.00 \\
capsules & 98.28$\rightarrow$98.83 & 98.83$\rightarrow$99.19 & 97.03$\rightarrow$97.51 & 99.60$\rightarrow$99.61 & 64.05$\rightarrow$67.34 & 63.54$\rightarrow$64.32 & 97.60$\rightarrow$98.29 \\
cashew & 98.48$\rightarrow$98.48 & 99.36$\rightarrow$99.35 & 97.49$\rightarrow$96.48 & 97.91$\rightarrow$96.36 & 64.18$\rightarrow$61.39 & 62.17$\rightarrow$61.15 & 94.81$\rightarrow$93.57 \\
chewinggum & 99.86$\rightarrow$99.86 & 99.93$\rightarrow$99.93 & 99.00$\rightarrow$98.52 & 99.28$\rightarrow$99.20 & 64.59$\rightarrow$75.91 & 66.86$\rightarrow$70.49 & 90.79$\rightarrow$89.38 \\
fryum & 99.42$\rightarrow$99.36 & 99.71$\rightarrow$99.69 & 97.54$\rightarrow$97.51 & 96.57$\rightarrow$96.15 & 48.57$\rightarrow$47.31 & 52.05$\rightarrow$50.61 & 93.26$\rightarrow$93.58 \\
macaroni1 & 98.45$\rightarrow$98.62 & 98.38$\rightarrow$98.54 & 94.74$\rightarrow$95.57 & 99.69$\rightarrow$99.71 & 33.49$\rightarrow$37.50 & 39.81$\rightarrow$41.72 & 97.09$\rightarrow$97.37 \\
macaroni2 & 97.73$\rightarrow$97.48 & 97.78$\rightarrow$97.60 & 93.66$\rightarrow$91.51 & 99.80$\rightarrow$99.80 & 25.89$\rightarrow$28.04 & 36.86$\rightarrow$37.93 & 98.78$\rightarrow$98.83 \\
pcb1 & 98.97$\rightarrow$99.11 & 98.91$\rightarrow$99.08 & 96.55$\rightarrow$96.55 & 99.64$\rightarrow$99.61 & 85.71$\rightarrow$86.48 & 79.10$\rightarrow$79.76 & 95.67$\rightarrow$95.87 \\
pcb2 & 98.74$\rightarrow$99.12 & 98.17$\rightarrow$98.88 & 98.02$\rightarrow$99.01 & 98.79$\rightarrow$98.75 & 36.43$\rightarrow$38.91 & 43.20$\rightarrow$44.40 & 90.79$\rightarrow$90.67 \\
pcb3 & 99.39$\rightarrow$99.33 & 99.42$\rightarrow$99.37 & 96.55$\rightarrow$96.55 & 99.12$\rightarrow$99.08 & 35.62$\rightarrow$33.20 & 43.33$\rightarrow$43.10 & 94.96$\rightarrow$94.18 \\
pcb4 & 99.91$\rightarrow$99.89 & 99.91$\rightarrow$99.89 & 98.52$\rightarrow$98.49 & 98.78$\rightarrow$98.78 & 50.73$\rightarrow$52.04 & 52.83$\rightarrow$53.22 & 93.74$\rightarrow$94.61 \\
pipe\_fryum & 99.76$\rightarrow$99.64 & 99.89$\rightarrow$99.84 & 99.00$\rightarrow$99.00 & 99.05$\rightarrow$99.14 & 55.89$\rightarrow$61.78 & 62.88$\rightarrow$65.01 & 94.20$\rightarrow$96.49 \\
\bottomrule
\end{tabular}}
\end{table}

\begin{table}[!htbp]
\centering
\scriptsize
\setlength{\tabcolsep}{3pt}
\caption{Category-level seed-0 results for Real-IAD Reconstruction. Cells with arrows report baseline $\rightarrow$ BoundarySupport.}
\resizebox{\textwidth}{!}{%
\begin{tabular}{lrrrrrrr}
\toprule
Category & I-AUROC & I-AP & I-F1 & P-AUROC & P-AP & P-F1 & AUPRO \\
\midrule
audiojack & 94.22$\rightarrow$93.78 & 91.73$\rightarrow$91.35 & 82.38$\rightarrow$80.92 & 99.81$\rightarrow$99.82 & 62.21$\rightarrow$66.55 & 61.46$\rightarrow$64.26 & 98.16$\rightarrow$98.31 \\
bottle\_cap & 95.78$\rightarrow$95.49 & 95.75$\rightarrow$95.34 & 88.93$\rightarrow$87.78 & 99.56$\rightarrow$99.57 & 34.92$\rightarrow$38.60 & 38.69$\rightarrow$41.58 & 97.14$\rightarrow$96.95 \\
button\_battery & 90.96$\rightarrow$89.80 & 92.74$\rightarrow$92.19 & 87.23$\rightarrow$86.29 & 99.50$\rightarrow$99.54 & 54.89$\rightarrow$62.53 & 65.91$\rightarrow$67.10 & 96.58$\rightarrow$96.36 \\
end\_cap & 87.06$\rightarrow$84.72 & 87.12$\rightarrow$84.88 & 85.65$\rightarrow$84.36 & 99.29$\rightarrow$99.28 & 17.66$\rightarrow$17.46 & 26.22$\rightarrow$26.72 & 96.86$\rightarrow$97.17 \\
eraser & 93.78$\rightarrow$93.72 & 93.00$\rightarrow$92.74 & 85.28$\rightarrow$84.72 & 99.78$\rightarrow$99.82 & 51.88$\rightarrow$58.67 & 53.61$\rightarrow$58.13 & 98.05$\rightarrow$98.23 \\
fire\_hood & 95.96$\rightarrow$94.89 & 92.61$\rightarrow$89.27 & 87.32$\rightarrow$84.59 & 99.77$\rightarrow$99.78 & 57.42$\rightarrow$63.05 & 58.13$\rightarrow$59.93 & 98.09$\rightarrow$97.88 \\
mint & 78.32$\rightarrow$77.79 & 85.63$\rightarrow$85.08 & 78.59$\rightarrow$77.98 & 98.95$\rightarrow$98.96 & 39.98$\rightarrow$43.87 & 45.08$\rightarrow$46.53 & 87.07$\rightarrow$88.28 \\
mounts & 97.49$\rightarrow$97.53 & 96.21$\rightarrow$96.24 & 89.30$\rightarrow$89.42 & 99.80$\rightarrow$99.84 & 52.17$\rightarrow$59.46 & 51.01$\rightarrow$56.47 & 98.54$\rightarrow$98.84 \\
pcb & 96.65$\rightarrow$96.85 & 98.04$\rightarrow$98.10 & 93.35$\rightarrow$93.60 & 99.80$\rightarrow$99.82 & 59.77$\rightarrow$63.02 & 59.10$\rightarrow$61.03 & 98.27$\rightarrow$98.41 \\
phone\_battery & 96.32$\rightarrow$95.89 & 96.78$\rightarrow$96.35 & 90.63$\rightarrow$89.56 & 99.85$\rightarrow$99.86 & 61.16$\rightarrow$66.35 & 57.60$\rightarrow$61.10 & 98.52$\rightarrow$98.67 \\
plastic\_nut & 95.27$\rightarrow$95.27 & 87.90$\rightarrow$87.58 & 84.62$\rightarrow$82.67 & 99.79$\rightarrow$99.78 & 47.37$\rightarrow$50.52 & 47.22$\rightarrow$49.72 & 97.66$\rightarrow$97.68 \\
plastic\_plug & 96.70$\rightarrow$95.95 & 95.76$\rightarrow$94.63 & 90.08$\rightarrow$88.21 & 99.71$\rightarrow$99.74 & 34.32$\rightarrow$43.80 & 40.97$\rightarrow$46.72 & 98.66$\rightarrow$98.60 \\
porcelain\_doll & 95.47$\rightarrow$95.28 & 92.37$\rightarrow$92.01 & 86.01$\rightarrow$85.71 & 99.71$\rightarrow$99.68 & 33.36$\rightarrow$37.59 & 38.86$\rightarrow$41.32 & 98.92$\rightarrow$98.84 \\
regulator & 96.70$\rightarrow$96.30 & 88.19$\rightarrow$89.48 & 83.58$\rightarrow$84.75 & 99.94$\rightarrow$99.93 & 55.60$\rightarrow$59.96 & 57.56$\rightarrow$59.30 & 99.58$\rightarrow$99.58 \\
rolled\_strip\_base & 99.71$\rightarrow$99.66 & 99.87$\rightarrow$99.85 & 98.83$\rightarrow$98.93 & 99.76$\rightarrow$99.77 & 50.65$\rightarrow$47.72 & 50.68$\rightarrow$50.55 & 98.78$\rightarrow$98.84 \\
sim\_card\_set & 99.17$\rightarrow$98.91 & 99.38$\rightarrow$99.17 & 96.60$\rightarrow$95.74 & 99.78$\rightarrow$99.81 & 64.71$\rightarrow$68.84 & 61.18$\rightarrow$63.38 & 97.99$\rightarrow$98.28 \\
switch & 99.11$\rightarrow$99.00 & 99.52$\rightarrow$99.46 & 96.62$\rightarrow$96.35 & 99.55$\rightarrow$99.29 & 63.06$\rightarrow$65.66 & 62.96$\rightarrow$63.36 & 98.38$\rightarrow$97.80 \\
tape & 99.58$\rightarrow$99.63 & 99.33$\rightarrow$99.42 & 96.63$\rightarrow$96.88 & 99.87$\rightarrow$99.90 & 46.09$\rightarrow$52.29 & 51.06$\rightarrow$56.29 & 99.47$\rightarrow$99.53 \\
terminalblock & 99.26$\rightarrow$99.15 & 99.54$\rightarrow$99.48 & 98.07$\rightarrow$97.94 & 99.89$\rightarrow$99.90 & 60.27$\rightarrow$63.63 & 60.30$\rightarrow$62.94 & 99.32$\rightarrow$99.41 \\
toothbrush & 91.41$\rightarrow$91.03 & 95.48$\rightarrow$95.27 & 87.89$\rightarrow$86.47 & 99.03$\rightarrow$99.04 & 48.64$\rightarrow$49.33 & 53.39$\rightarrow$53.95 & 94.69$\rightarrow$95.06 \\
toy & 88.94$\rightarrow$88.12 & 94.96$\rightarrow$94.43 & 86.17$\rightarrow$85.58 & 92.28$\rightarrow$91.21 & 23.18$\rightarrow$24.24 & 33.36$\rightarrow$35.01 & 87.59$\rightarrow$86.17 \\
toy\_brick & 82.53$\rightarrow$79.90 & 83.28$\rightarrow$79.00 & 73.80$\rightarrow$71.40 & 97.92$\rightarrow$97.93 & 48.33$\rightarrow$54.91 & 53.59$\rightarrow$56.43 & 82.54$\rightarrow$84.16 \\
transistor1 & 96.67$\rightarrow$96.81 & 98.03$\rightarrow$98.08 & 93.42$\rightarrow$94.01 & 99.38$\rightarrow$99.42 & 48.90$\rightarrow$52.34 & 49.40$\rightarrow$52.25 & 97.10$\rightarrow$97.53 \\
u\_block & 96.49$\rightarrow$96.47 & 92.62$\rightarrow$92.22 & 86.89$\rightarrow$85.95 & 99.80$\rightarrow$99.81 & 43.79$\rightarrow$48.91 & 47.50$\rightarrow$50.80 & 97.80$\rightarrow$98.03 \\
usb & 96.92$\rightarrow$96.92 & 96.35$\rightarrow$96.35 & 93.04$\rightarrow$92.75 & 99.83$\rightarrow$99.84 & 52.28$\rightarrow$54.60 & 54.05$\rightarrow$55.55 & 99.13$\rightarrow$99.12 \\
usb\_adaptor & 93.04$\rightarrow$92.48 & 90.68$\rightarrow$90.23 & 84.42$\rightarrow$83.13 & 99.72$\rightarrow$99.75 & 29.59$\rightarrow$33.06 & 35.63$\rightarrow$38.72 & 97.73$\rightarrow$98.27 \\
vcpill & 94.64$\rightarrow$94.36 & 93.19$\rightarrow$93.10 & 85.01$\rightarrow$85.29 & 98.93$\rightarrow$98.96 & 73.67$\rightarrow$77.06 & 71.21$\rightarrow$73.55 & 91.85$\rightarrow$91.83 \\
wooden\_beads & 88.80$\rightarrow$87.52 & 92.19$\rightarrow$90.81 & 82.92$\rightarrow$81.78 & 98.90$\rightarrow$98.91 & 46.64$\rightarrow$50.52 & 50.16$\rightarrow$53.42 & 90.71$\rightarrow$92.07 \\
woodstick & 91.37$\rightarrow$89.47 & 78.33$\rightarrow$74.19 & 73.04$\rightarrow$66.97 & 99.58$\rightarrow$99.58 & 56.81$\rightarrow$65.52 & 58.70$\rightarrow$62.92 & 93.93$\rightarrow$93.80 \\
zipper & 99.71$\rightarrow$99.64 & 99.85$\rightarrow$99.82 & 98.11$\rightarrow$98.10 & 99.16$\rightarrow$99.13 & 64.53$\rightarrow$69.73 & 67.21$\rightarrow$69.66 & 98.04$\rightarrow$98.19 \\
\bottomrule
\end{tabular}}
\end{table}

\FloatBarrier

\FloatBarrier
\section{Score Changes and Failure Cases}
\label{app:scores}
\subsection{Reconstruction score changes}
\begin{table}[h]
\centering
\small
\caption{Seed-0 reconstruction score changes.}
\begin{tabular}{lrrrr}
\toprule
Dataset & $\Delta$normal q95 & $\Delta$normal q99 & $\Delta$defect mean & $\Delta$P-AP \\
\midrule
MVTec & -1.91 & -4.91 & -4.01 & +4.94 \\
VisA & -0.28 & -0.88 & -1.29 & +2.09 \\
Real-IAD & -0.32 & -1.13 & -2.49 & +4.20 \\
\bottomrule
\end{tabular}
\end{table}
At seed 0, high anomaly scores on normal regions and mean defect scores decrease together on all three datasets while P-AP increases. The same macro direction repeats over all completed MVTec and VisA seeds.

\subsection{Memory score changes}
For memory, we also report the excess normal score
\[
e_q=Q_q(\text{normal patches in defective images})-Q_q(\text{patches in clean images}).
\]

\begin{table}[h]
\centering
\small
\caption{Seed-0 memory score changes under the uniform headline policies.}
\begin{tabular}{lrrrrrr}
\toprule
Dataset & Abs. q95 & Abs. q99 & Excess q95 & Excess q99 & Defect mean & $\Delta$P-AP \\
\midrule
MVTec & -0.35 & -1.02 & -0.51 & -1.14 & -0.51 & +2.26 \\
VisA & +0.30 & -0.43 & -0.33 & -0.92 & -0.52 & +1.82 \\
Real-IAD & +0.55 & +0.09 & -0.11 & -0.48 & -0.21 & +1.76 \\
\bottomrule
\end{tabular}
\end{table}
Absolute normal scores do not move in one direction on every dataset. The defect-image-minus-clean-image excess q95/q99 decreases in all three settings.

\subsection{Representative failure cases}
\begin{table}[!htbp]
\centering
\small
\caption{Selected seed-0 negative and metric-trade-off cases.}
\begin{tabular}{p{0.36\textwidth}p{0.58\textwidth}}
\toprule
Case & Observed change \\
\midrule
Reconstruction / VisA / cashew & P-AP -2.79; P-AUROC -1.56; q99 -0.74; defect mean -1.79. \\
Reconstruction / VisA / pcb3 & P-AP -2.42; AUPRO -0.78. \\
Reconstruction / VisA / fryum & P-AP -1.26; AUPRO +0.32. \\
Reconstruction / Real-IAD / rolled strip base & P-AP -2.92 while P-AUROC and AUPRO increase slightly. \\
Reconstruction / Real-IAD / end cap & P-AP -0.20; I-AUROC -2.33. \\
Memory / Real-IAD / u block & P-AP +3.32 while absolute q95 and q99 increase. \\
\bottomrule
\end{tabular}
\end{table}
At seed 0, P-AP is positive in 57/57 memory categories and 52/57 reconstruction categories. Across all three seeds, the corresponding sign counts are 171/171 and 158/171. The clearest aggregate trade-off is Real-IAD reconstruction: P-AP/P-F1/AUPRO change by $+4.20\pm0.12$, $+2.56\pm0.09$, and $+0.16\pm0.03$, while I-AUROC changes by $-0.52\pm0.04$.

\section{Uniform VisA Memory Comparison}
\label{app:visa_policy}
The final VisA memory result applies one uniform q50/q90 rule to all 12 categories and all three seeds. A previously explored mixed policy was evaluated at seed 0 and reached 54.12 P-AP, compared with 54.05 for the uniform seed-0 policy and 52.23 for the clean baseline. The 0.07-point difference is small relative to the gain over the clean memory; the main paper therefore uses the uniform policy.

\section{Compute and Reproducibility Details}
\label{app:compute}
Memory feature caches and banks are stored in FP16, while candidate, coreset, and readout distances are computed in FP32. Reconstruction uses FP32 training, 5,000 iterations per category, and batch size 16.

\begin{table}[h]
\centering
\small
\caption{Seed-0 reconstruction category-GPU hours. Totals are sums over categories, not parallel wall-clock time.}
\begin{tabular}{lrrr}
\toprule
Dataset & Baseline hours & BoundarySupport hours & Ratio \\
\midrule
MVTec & 5.81 & 15.17 & 2.61$\times$ \\
VisA & 4.99 & 12.71 & 2.55$\times$ \\
Real-IAD & 12.60 & 31.21 & 2.48$\times$ \\
\bottomrule
\end{tabular}
\end{table}
BoundarySupport reconstruction increases training cost in these runs while leaving the deployed architecture and scoring path unchanged. The memory branch likewise keeps the final bank size and readout fixed; its added work occurs during offline synthesis, candidate scoring, and bank reconstruction.

The release artifact will include category-level result files, fixed-budget and distance-diagnostic aggregation, intervention and mask construction, memory candidate selection, reconstruction target controls, and table/figure regeneration scripts. Reported tables are generated from category-level results using ROUND\_HALF\_UP to two decimals; algorithm-defining constants retain their stated precision.

\FloatBarrier
\clearpage
\section{Full 57-Category Qualitative Visualization}
\label{app:allqual}
The following pages provide an all-category qualitative comparison across MVTec AD, VisA, and Real-IAD. Each row follows the same column order: input image, ground-truth mask, ProCon, ProCon + BoundarySupport, Dinomaly, and Dinomaly + BoundarySupport. The quantitative category-level results in Appendix~\ref{app:categories} remain the basis for performance claims; these figures show the corresponding spatial predictions across all 57 benchmark categories.

\paragraph{MVTec AD.}
\begin{center}
\includegraphics[width=\textwidth,page=1]{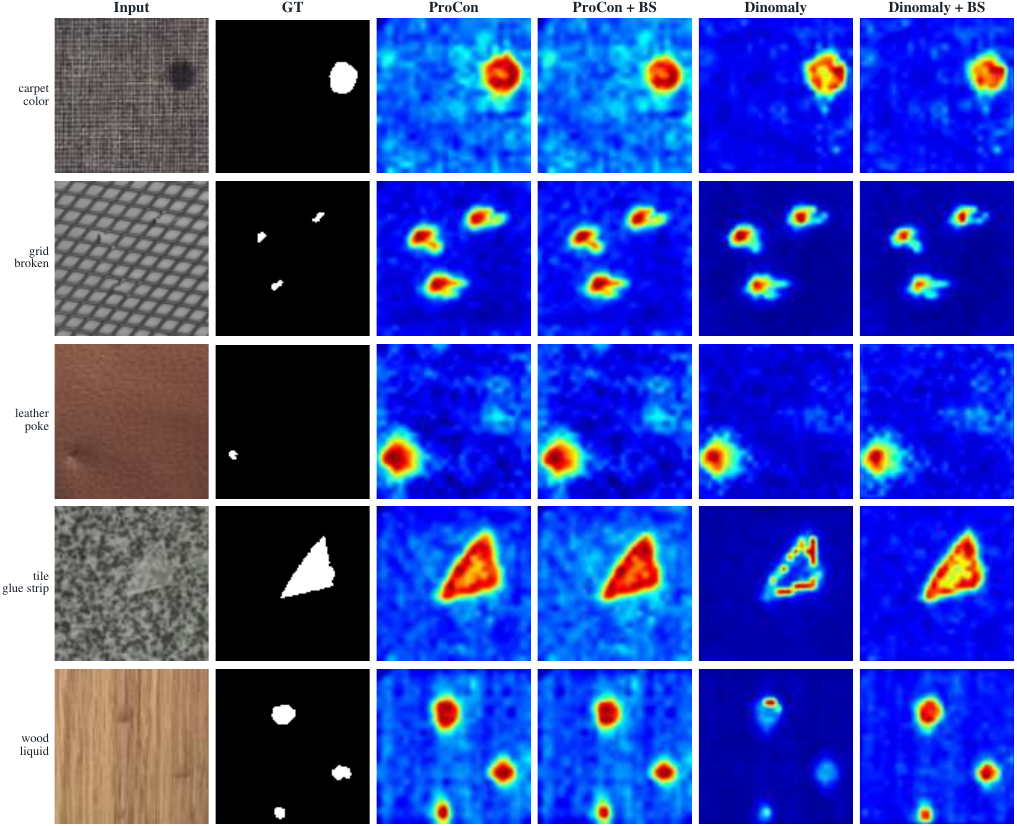}
\end{center}
\clearpage
\begin{center}
\includegraphics[width=\textwidth,page=2]{figures/all_category_JET_57_categories.pdf}
\end{center}
\clearpage
\begin{center}
\includegraphics[width=\textwidth,page=3]{figures/all_category_JET_57_categories.pdf}
\end{center}
\clearpage

\paragraph{VisA.}
\begin{center}
\includegraphics[width=\textwidth,page=4]{figures/all_category_JET_57_categories.pdf}
\end{center}
\clearpage
\begin{center}
\includegraphics[width=\textwidth,page=5]{figures/all_category_JET_57_categories.pdf}
\end{center}
\clearpage
\begin{center}
\includegraphics[width=\textwidth,page=6]{figures/all_category_JET_57_categories.pdf}
\end{center}
\clearpage

\paragraph{Real-IAD.}
\begin{center}
\includegraphics[width=\textwidth,page=7]{figures/all_category_JET_57_categories.pdf}
\end{center}
\clearpage
\begin{center}
\includegraphics[width=\textwidth,page=8]{figures/all_category_JET_57_categories.pdf}
\end{center}
\clearpage
\begin{center}
\includegraphics[width=\textwidth,page=9]{figures/all_category_JET_57_categories.pdf}
\end{center}
\clearpage
\begin{center}
\includegraphics[width=\textwidth,page=10]{figures/all_category_JET_57_categories.pdf}
\end{center}
\clearpage
\begin{center}
\includegraphics[width=\textwidth,page=11]{figures/all_category_JET_57_categories.pdf}
\end{center}
\clearpage
\begin{center}
\includegraphics[width=\textwidth,page=12]{figures/all_category_JET_57_categories.pdf}
\end{center}
\clearpage

\end{document}